%% file: acl2019.tex
\documentclass[11pt,a4paper]{article}
\usepackage[hyperref]{acl2021}
\usepackage{times}
\usepackage{latexsym}
\usepackage{amsmath}
\usepackage{url}
\usepackage{graphicx}
\usepackage{multirow}
\usepackage{booktabs}
\usepackage{amsfonts}
\usepackage{verbatim}
\usepackage{hyperref}
\usepackage{enumitem}
\usepackage{multirow}
\usepackage{wrapfig}
\usepackage{subfigure}
\usepackage{amssymb,mathrsfs}
\usepackage{natbib}

\usepackage[normalem]{ulem}

\usepackage{microtype}

\usepackage{tensor}
\aclfinalcopy % Uncomment this line for the final submission
\def\aclpaperid{2079} %  Enter the acl Paper ID here

\title{Good for Misconceived Reasons: An Empirical Revisiting on the Need for Visual Context in Multimodal Machine Translation}

\author{
 Zhiyong Wu$^{\dag}$~\thanks{\, The majority of this work was done while the first author was interning at Tencent AI Lab.}\,, Lingpeng Kong$^{\dag\P}$, Wei Bi$^{\ddag}$, Xiang Li$^{\S}$, Ben Kao$^{\dag}$\\
 $^{\dag}$The University of Hong Kong, \\
 $^{\ddag}$Tencent AI Lab, $^\P$Shanghai Artificial Intelligence Laboratory, $^{\S}$East China Normal University \\
 $^{\dag}$\{zywu,lpk,kao\}@cs.hku.hk, $^{\ddag}$victoriabi@tencent.com, $^{\S}$xiangli@dase.ecnu.edu.cn\\
}
\definecolor{RED}{RGB}{228, 26, 28}
\definecolor{BLUE}{RGB}{55, 126, 184}
\definecolor{GREEN}{RGB}{77, 175, 74}
\newcommand{\model}{RMMT}

\usepackage{amsmath,amsfonts,bm}
\usepackage{float}
\usepackage{ulem}
\newcommand{\dataset}{Multi30k}
\date{}

\begin{document}
\maketitle
\begin{abstract}
A neural multimodal machine translation (MMT) system is one that aims to perform better translation by extending conventional text-only translation models with multimodal information. Many recent studies report improvements when equipping their models with the multimodal module, despite the controversy of whether such improvements indeed come from the multimodal part. We revisit the contribution of multimodal information in MMT by devising two \textit{interpretable} MMT models. To our surprise, although our models replicate similar gains as recently developed multimodal-integrated systems achieved, 
our models learn to \textit{ignore} the multimodal information. Upon further investigation, we discover that the improvements achieved by the multimodal models over text-only counterparts are in fact results of the regularization effect. 
We report empirical findings that highlight the importance of MMT models' interpretability, and discuss how our findings will benefit future research.
%set new paradigms for future MMT research.

\end{abstract}

\input{1-intro}

\input{2-related}

\input{3-method}

\input{4-effectiveness}
\input{5-analysis}

\section{Conclusion}
In this paper we devise two interpretable models that exhibit state-of-the-art performance on the widely adopted MMT datasets --- \dataset\ and the new video-based dataset --- VaTex. Our analysis on the proposed models, as well as on other existing MMT systems, suggests that visual context helps MMT in the similar vein as regularization methods (e.g., weight decay), under sufficient textual context. Those empirical findings, however, should not be understood as us downplaying the importance existing datasets and models; we believe that sophisticated MMT models are necessary for effective grounding of visual context into translation. Our goal, rather, is to (1) provide additional clarity on the \textit{remaining} shortcomings of current dataset and stress the need for new datasets to move the field forward; (2) emphasise the importance of interpretability in MMT research.
%denying the contribution of existing MMT methods.  Their effectiveness might be impaired by limitation in existing datasets~\citep{caglayan2019probing}. 
\section{Acknowledgement}
Zhiyong Wu is partially supported by a research grant from the HKU-TCL Joint Research Centre for Artificial Intelligence.

\bibliographystyle{acl_natbib}
\bibliography{acl2019}
\input{9-appendix}

\end{document}

%% file: 1-intro.tex
\section{Introduction}
%It has long been researchers' belief that with appropriate regularization, adding more features to NLP models rarely harms performance. However, 
%With the recent prevalence of \textit{deep} neural network models~\citep{Goodfellow-et-al-2016,bishop1995training}, the term ``regularization'' is no longer reserved solely for a penalty term in the loss function (Bishop, 1995a). Instead, regularization can be any supplementary technique or input that aims at making the model generalize better~\citep(kukavcka2017regularization). Such as dropout, Gaussian noise, batch normalization, or even image/sound transformations.

Multimodal Machine Translation (MMT) aims at designing better translation systems by extending conventional text-only translation systems to take into account multimodal information, especially from visual modality~\citep{mmt2016,vatex}.  
Despite many previous success in MMT that report improvements when models are equipped with visual information~\citep{calixto-etal-2017-doubly,helcl2018doubly,ive-etal-2019-distilling,lin2020dynamic,yin2020novel}, 
there have been continuing debates on the need for visual context in MMT.

In particular, \citet{mmt2016,mmt2017,mmt2018} argue that visual context does not seem to help translation reliably, at least as measured by automatic metrics. \citet{elliott2018adversarial, gronroos-etal-2018-memad} provide further evidence by showing that MMT models are, in fact, insensitive to visual input and can translate without significant performance losses even in the presence of features derived from unrelated images. A more recent study~\citep{caglayan2019probing}, however, shows that under limited textual context (e.g., noun words are masked), models 
can leverage visual input to generate better translations. But it remains unclear where the gains of MMT methods come from, when the textual context is complete. 
%Despite those concerns, the field appears to be progressing steadily, albeit slowly in recent years. 

The main tool utilized in prior discussion is \textit{adversarial model comparison} ---  explaining the behavior of complex and black-box MMT models by comparing performance changes when given adversarial input (e.g., random images). Although such an opaque tool is an acceptable beginning to investigate the need for visual context in MMT, they  provide rather indirect evidence~\cite{hessel2020does}. This is because performance differences can often be attributed to factors unrelated to visual input, such as %hyperparameter search schemes~\cite{yogatama2015bayesian}, 
regularization~\cite{kukavcka2017regularization}, data bias~\cite{jabri2016revisiting}, and some others~\cite{dodge-etal-2019-show}. 

From these perspectives, we revisit the need for visual context in MMT by designing two interpretable models. Instead of directly infusing visual features into the model, we design learnable components, which allow the model to voluntarily decide the usefulness of the visual features and reinforce their effects when they are helpful. 
To our surprise, while our models are shown to be
effective
on \dataset~\citep{elliott2016multi30k} and VaTex~\citep{vatex} datasets,
they learn to \textit{ignore} the multimodal information. %\li{despite demonstrated effectiveness on \dataset~\citep{elliott2016multi30k} and VaTex~\citep{vatex} dataset}. 
Our further analysis suggests that under sufficient textual context, the improvements come from a regularization effect that is similar to random noise injection~\citep{bishop1995training} and weight decay~\citep{hanson1989comparing}. The additional visual information is treated as noise signals that can be used to
enhance model training and lead to a more robust network with 
lower generalization error~\citep{salamon2017deep}.
%lower the model's trust in the hidden representations generated from the neural networks, resulting in a more robust network with lower generalization error~\citep{salamon2017deep}. 
Repeating the evaluation under limited textual context further substantiates our findings and complements previous analysis~\citep{caglayan2019probing}. 

Our contributions are twofold. 
First, we revisit the need for visual context in the popular task of multimodal machine translation and find that: (1) under sufficient textual context, the MMT models' improvements over text-only counterparts result from the regularization effect (Section~\ref{sec:revisit}). (2) under limited textual context, MMT models can leverage visual context to help translation (Section~\ref{sec:limited}). Our findings highlight the importance of MMT models' interpretability and the need for a new benchmark to advance the community.
%and the need for a new benchmark to move the community forward. 

Second, for the MMT task, we provide a strong text-only baseline implementation and two models with interpretable components that replicate similar gains as reported in previous works. Different from adversarial model comparison methods, our models are interpretable due to the specifically designed model structure and can serve as standard baselines for future interpretable MMT studies. Our code is available at \url{https://github.com/LividWo/Revisit-MMT}.

%% file: 2-related.tex
\section{Background}
\label{sec:related}
%\bi{need to reorganize. need to add: 1 other visual-grounding text generation task; 2 interpretable models on various generation task. }
One can broadly categorize MMT systems into two types: (1) \underline{Conventional} MMT, where there is gold alignment between the source (target) sentence pair and a relevant image and (2) \underline{Retrieval-based} MMT, where systems retrieve relevant images from an image corpus as additional clues to assist translation. 

\paragraph{Conventional MMT}
% We review recent progress in MMT with respect to images' entry point in the pipeline, namely, input sentence, encoder, decoder and output sentence.
Most MMT systems 
%follow the conventional settings and 
require datasets consist of images with bilingual annotations for both training and inference. 
% multimodal information that is provided along with the bilingual sentence pairs in the dataset.
Many early attempts use a pre-trained model (e.g., ResNet~\citep{he2016deep}) to encode images into feature vectors. 
This visual representation can be used to initialize the encoder/decoder's hidden vectors~\citep{elliott2015multilingual,libovicky-helcl-2017-attention,calixto2016dcu}.
It can also be appended/prepended to word embeddings as additional input tokens~\citep{huang-etal-2016-attention,calixto-liu-2017-sentence}.
Recent works~\citep{libovicky-etal-2018-input,zhou2018visual,ive-etal-2019-distilling, lin2020dynamic} employ attention mechanism to generate a visual-aware representation for the decoder. For instance, \textit{Doubly-ATT}~\citep{calixto-etal-2017-doubly, helcl2018doubly, arslan2018doubly} insert an extra visual attention sub-layer between the decoder's source-target attention sub-layer and feed-forward sub-layer. 
While there are more works on engineering decoders,
encoder-based approaches are relatively less explored. To this end, \citet{yao2020multimodal} and \citet{yin2020novel} replace the vanilla Transformer encoder with a  multi-modal encoder. 

Besides the exploration on network structure, researchers also propose to leverage the benefits of multi-tasking to improve MMT~\cite{elliott2017imagination,zhou2018visual}. The \textit{Imagination} architecture~\cite{elliott2017imagination, helcl2018doubly} decomposes multimodal translation into two sub-tasks: translation task and an auxiliary visual reconstruction task, which encourages the model to learn a visually grounded source sentence representation.

\paragraph{Retrieval-based MMT}
The effectiveness of conventional MMT heavily relies on the availability of images with bilingual annotations. 
This could restrict its wide applicability. 
To address this issue, \citet{zhang2020neural} propose \textit{UVR-NMT} that integrates a retrieval component into MMT. They use TF-IDF to build a token-to-image lookup table, 
based on which images sharing similar topics with a source sentence are retrieved as relevant images. This creates image-bilingual-annotation instances for training. 
Retrieval-based models have been shown to improve performance across a variety of NLP tasks besides MMT, such as question answering~\citep{guu2020realm}, %fact checking~\citep{thorne2018fever}, 
dialogue~\citep{weston2018retrieve}, 
language modeling~\citep{khandelwal2019generalization}, question generation~\citep{lewis2020retrieval}, and translation~\citep{gu2018search}.

%% file: 3-method.tex
\section{Method}

In this section we introduce two interpretable MMT models: (1) \textit{Gated Fusion} for conventional MMT and (2) \textit{Dense-Retrieval-augmented MMT} (\model) for retrieval-based MMT. 
Our design philosophy is that models should learn, in an interpretable manner, to which degree multimodal information is used. 
%Following this principle, we focus on the component that is responsible for the integration of multimodal information.
Following this principle, we focus on the component that integrates multimodal information.
In particular, %we use a gating matrix $\Lambda$ ~\cite{yin2020novel, zhang2020neural} to control the amount of visual information that will be blended into the textual representation 
we use a gating matrix $\Lambda$ ~\cite{yin2020novel, zhang2020neural} to control the amount of visual information to be blended into the textual representation.  
%The learned gating matrix $\Lambda$ 
Such a matrix
%makes the fusion process interpretable 
facilitates interpreting the fusion process: a larger gating value $\Lambda_{ij} \in [0, 1]$ indicates that the model exploits more visual context in translation, and vice versa.

%The fundamental component in a MMT system is investigating how multimodal information should be integrated into the translation model 
% \ben{Remove?: rather than directly adding it into the model as features}.
%The core of our designs is a gating vector $\Vec{\lambda} \in[0,1]^d$ that controls the amount of visual information that is blended into the textual representation. The learned $\Vec{\lambda}$ makes the fusion process interpretable: a larger gating value $\lambda \in \Vec{\lambda}$ indicates that the model exploits more visual context in the  translation.
%We start with the simple yet effective \textit{Gated Fusion} model that translates a source sentence $x$ into a target $y$ with the help of a gold-standard image $z$ associated to $x$ and $y$. The model learns a gating vector $\Vec{\lambda} \in[0,1]^d$ that controls the amount of visual information that is blended into the textual representation. The learned $\Vec{\lambda}$ makes the fusion process interpretable: a larger gating value $\lambda \in \Vec{\lambda}$ indicates that the model exploits more visual context in the  translation.

%For \model, we augment a conventional MMT model with a learnable retriever, which learns to retrieve images that are semantically-relevant to an input sentence $x$. We jointly train the retriever with the rest of the model in an end-to-end manner such that the model \textit{learns to retrieve} images that are highly influential in the translation. 

\subsection{Gated Fusion MMT}
\label{sec:gated}
Given a source sentence $x$ of length $T$ and an associated image $z$, we compute the probability of generating target sentence $y$ of length $N$ by:

\begin{equation}
	\label{eq:gated}
	p(y|x, z) = \prod_{i}^{N} p_{\theta}\left(y_{i} \mid x, z, y_{<i}\right),
\end{equation}
where $p_{\theta}\left(y_{i} \mid x, z, y_{<i}\right)$ is implemented with
a Transformer-based~\citep{vaswani2017attention} network. 
Specifically, we first feed $x$ into a vanilla Transformer encoder to obtain a textual representation $\mathbf{H}_{\text{text}} \in \mathbb{R}^{T\times d}$, which is then fused with visual representation $\text { Embed }_{\text{image }}(z)$ before fed into the Transformer decoder. 
%The encoder contains $L$ identical layers, each with a multi-head self-attention sublayer followed by a position-wise, fully connected feed-forward sublayer. Layers and sublayers are chained using residual connection~\citep{he2016deep} and layer normalization~\citep{ba2016layer}. 
For each image $z$, we use a pre-trained ResNet-50 CNN~\citep{he2016deep} 
%trained on ImageNet~\citep{imagenet} 
to extract a 2048-dimensional average-pooled visual representation, which is then projected to the same dimension as $\mathbf{H}_{\text{text}}$:
\begin{equation}
\label{eq:resnet}
\text { Embed }_{\text{image }}(z)=\mathbf{W}_{\text {z }} \operatorname{ResNet}_{\mathrm{pool}}\left(z\right).
\end{equation} 

%The fusion of  $\mathbf{H}_{\text{text}}$ and $\text{ Embed }_{\text{image }}(z)$ is controlled by a gating matrix $\Lambda \in [0,1]^{T\times d}$
We next generate a gating matrix $\Lambda \in [0,1]^{T\times d}$ to control the fusion of $\mathbf{H}_{\text{text}}$ and $\text{ Embed }_{\text{image }}(z)$:
%We now devise a gating mechanism to control the fusion of $\mathbf{H}_{\text{text}}$ and $\text{ Embed }_{\text{image }}(z)$. The gating vector $\Vec{\lambda} \in [0,1]^d$ is computed as follow:
% \begin{equation}
$$
\Lambda =\operatorname{sigmoid}\left(\mathbf{W}_{\Lambda} \text{ Embed }_{\text{image }}(z)+\mathbf{U}_{\Lambda} \mathbf{H}_{\text{text}}\right),
% \end{equation}
$$
where $\mathbf{W}_{\Lambda}$ and $\mathbf{U}_{\Lambda}$ are model parameters. Note that this gating mechanism has been a building block for many recent MMT systems~\citep{zhang2020neural, lin2020dynamic, yin2020novel}. 
We are, however, the first to focus on its interpretability.
Finally, we generate the output vector $\mathbf{H}$ by:
% \lpk{if the model wants to use the image information, it will have to decrease the use of text information, which might not be the best case. $\mathbf{H}_\text{text} + \Vec{\lambda} \text{ Embed }_{\text{image }}(z)$ maybe? although i think in terms of results it won't matter.}
\begin{equation}
\label{eq:fusion}
\mathbf{H}=  \mathbf{H}_{\text{text}} + \Lambda \text{ Embed }_{\text{image }}(z).
\end{equation}
$\mathbf{H}$ is then fed into the decoder directly for translation as in vanilla Transformer.

\subsection{Retrieval-Augmented MMT (RMMT)}
\label{sec:retrieval}
\model\ consists of two sequential components: (1) an image retriever $p(z|x)$ that takes $x$ as input and returns Top-$K$ most relevant images from an image database; 
(2) a multi-modal translator $p(y|x, \mathcal{Z}) = \prod_{i}^{N} p_{\theta}\left(y_{i} \mid x, \mathcal{Z}, y_{<i}\right)$ that generates each $y_i$ conditioned on the input sentence $x$, the image set $\mathcal{Z}$ returned by the retriever, and the previously generated tokens $y_{<i}$. 
%Ideally, we can implement this multimodal translator using any existing MMT models. For the sake of interpretability, here we use the Gated Fusion model as an example. 

\paragraph{Image Retriever}
%\citet{zhang2020neural} built retriever using traditional sparse vector space models (TF-IDF), which is efficient, but usually the search is based on keywords and failed to consider full context of the query, resulting in pool retrieval performance.

Based on the TF-IDF model, searching in existing retrieval-based MMT~\cite{zhang2020neural} ignores the context information of a given query,
which could lead to poor performance.
To improve the recall of our image retriever, we compute the similarity between a sentence $x$ and an image $z$ with inner product:
% \begin{equation}
$$
%\label{eq:retriever} p(z \mid x) &=\frac{\exp f(x, z)}{\sum_{z^{\prime}} \exp f\left(x, z^{\prime}\right)}, \\ 
\label{eq:fxz} sim(x, z) =\text { Embed }_{\text {text }}(x)^{\top} \text { Embed }_{\operatorname{image}}(z), 
$$
where $\text{Embed}_{\text{text}}(x)$ and $\text{Embed}_{\text{image}}(z)$ are %functions that map $x$ and $z$ respectively to $d-$dimensional vectors.
$d$-dimensional representations of $x$ and $z$, respectively.
%The similarity score $sim(x, z)$ between $x$ and $z$ is defined as the inner product of their vector embeddings. 
%\li{remove the sentence.}
We then retrieve top-$K$ images that are closest to $x$. 
%The retrieval distribution is the softmax over all relevance scores. 
%We obtain $\text{Embed}_{\text{image}}(z)$ as in Equation~\ref{eq:resnet}.
For $\text{Embed}_{\text{image}}(z)$, we compute it by Eq.~\ref{eq:resnet}.
For $\text{Embed}_{\text{text}}(x)$, we implement it using BERT~\citep{devlin2019bert}:
\begin{equation}
\text { Embed }_{\text {text }}(x)=\mathbf{W}_{\text {text }} \operatorname{BERT}_{\mathrm{CLS}}\left(x\right).
\end{equation}
Following standard practices, we use a pre-trained BERT model\footnote{Here we use bert-base-uncased version.} % from HuggingFace~\citep{Wolf2019HuggingFacesTS}.
to obtain the ``pooled'' representation of the sequence (denoted as $\text{BERT}_{\mathrm{CLS}}(x)$). Here,
$\mathbf{W}_{\text {text }}$ is a projection matrix.

\paragraph{Multimodal Translator}
Different from Gated Fusion, $p(y|x, \mathcal{Z})$ now is conditioning on a set of images rather than one single image.  For each $z$ in $\mathcal{Z}$, we represent it using $\text{Embed}_{\text{image}}(z) \in \mathbb{R}^{d}$ as in Equation~\ref{eq:resnet}. 
%The core of \model\ is then to translate based on the image feature collections $\text{Embed}_{\text{image}}(\mathcal{Z}) \in \mathbb{R}^{K \times d}$, where $K=|\mathcal{Z}|$.
The image set $\mathcal{Z}$ then forms a feature matrix $\text{Embed}_{\text{image}}(\mathcal{Z}) \in \mathbb{R}^{K \times d}$, where $K=|\mathcal{Z}|$ and each row corresponds to the feature vector of an image.
We use a transformation layer $f_{\theta}(*)$ to extract salient features from $\text{Embed}_{\text{image}}(\mathcal{Z})$ and obtain a compressed representation $\mathbb{R}^{d}$ of $\mathcal{Z}$. 
After the transformation, ideally, we can implement $p(y|x, \mathcal{Z})$ using any existing MMT models. For interpretability, we follow the Gated Fusion model to fuse the textual and visual representations with a learnable gating matrix $\Lambda$:
\begin{equation}
\label{eq:rmmtfusion}
\mathbf{H}=  \mathbf{H}_{\text{text}} + \Lambda f_{\theta}(\text{ Embed }_{\text{image }}(\mathcal{Z})).
\end{equation}
%where $f_{\theta}(*)$ is a max-pooling layer with window size $K\times 1$.
Here, $f_{\theta}(*)$ denotes a max-pooling layer with window size $K\times 1$.

%% file: 4-effectiveness.tex
\section{Experiment}
In this section, we evaluate our models on the \dataset\ and VaTex benchmark. %and
%and VaTex~\cite{vatex} (See Appendix~\ref{app:vatex}).
%We report state-of-the-art results and confirm previous findings 
%replicate previous findings that multimodal context is helpful to translation. 
%We refer readers to Appendix for results on VaTex~\cite{vatex}, due to space limitation.
%\bi{how is the result and conclusion on VaTex? need to mention whether it has consistent results with multi30k. otherwise, it is useless to mention about it. also, need to highlight what this sec wants to tell the readers in the beginning.}
\subsection{Dataset}
\label{sec:dataset}
We perform experiments on the widely-used MMT datasets: \dataset.
%
%The \dataset\ dataset is an extension of Flickr30k~\citep{plummer2015flickr30k}. Flickr30k contains 31,014 images sourced from online photo-sharing websites. Each image is paired with five English descriptions. Researchers create \dataset\ by sampling one image-caption pair for each image from Flickr30k and crowd-sourcing German/French translation for the English caption. 
We follow a standard split of 29,000 instances for training, 1,014 for validation and 1,000 for testing (Test2016).
We also report results on the 2017 test set (Test2017) with extra 1,000 instances and the MSCOCO test set that includes
461 more challenging out-of-domain instances with ambiguous verbs. 
%
% The VaTex dataset contains 129,955 English-Chinese sentence pairs for training, 15,000 for validation, and 30,000 for testing. Each pair of sentence is associated with a video clip. In the test set, only the English sentences are provided, so we submit our  English$\rightarrow$Chinese translation to official website\footnote{https://competitions.codalab.org/competitions/24384} for evaluation.
%
%We use the official preprocessed version of \dataset\footnote{https://github.com/multi30k/dataset} and merge the source and target sentences to build a joint vocabulary. 
We merge the source and target sentences in the officially preprocessed version of \dataset\footnote{https://github.com/multi30k/dataset} to build a joint vocabulary.
%We then apply the byte pair encoding (BPE) algorithm~\citep{sennrich2016subword} with 10,000 merging operations to segment words into subwords, following standard practice, resulting in a vocabulary of 9,712 (9,544) tokens for En-De (En-Fr).
We then apply the byte pair encoding (BPE) algorithm~\citep{sennrich2016subword} with 10,000 merging operations to segment words into subwords, 
which generates a vocabulary of 9,712 (9,544) tokens for En-De (En-Fr).

\noindent
\textbf{Retriever pre-training.} We pre-train the retriever on a subset of the Flickr30k dataset~\citep{plummer2015flickr30k} that has overlapping instances with \dataset\ removed. We use \dataset's validation set to evaluate the retriever. %We measure the performance by recall-at-$K$ ($R@K$) defined as the fraction of queries for which the correct images are retrieved in the closest $K$ images to the query.
We measure the performance by recall-at-$K$ ($R@K$),
which is defined as the fraction of queries 
whose closest $K$ images retrieved contain the correct images.
The pre-trained retriever achieves $R@1$ of $22.8\%$ and $R@5$ of $39.6\%$.

%     Model    & Acc@1 & Acc@5 & Acc@10 & Acc@50 \\ \hline
%     Retriever& 22.8  & 39.6  & 46.3   & 66.3 \\

\subsection{Setup}
%We implement the proposed models and baselines with FairSeq~\citep{ott2019fairseq}. %We experiment with different model sizes (\textit{Base}, \textit{Small}, and \textit{Tiny}) as detailed in Appendix~\ref{app:size} along with training settings 
We experiment with different model sizes (\textit{Base}, \textit{Small}, and \textit{Tiny}, see Appendix~\ref{app:size} for details).
\textit{Base} is a widely-used model configuration for Transformer in both text-only translation~\citep{vaswani2017attention} and MMT~\citep{gronroos2018memad, ive-etal-2019-distilling}. However, for small datasets like \dataset, training such a large model (about 50 million parameters) could cause overfitting. In our preliminary study, we found that even a \textit{Small} configuration, which is commonly used for low-resourced translation~\citep{zhu2019incorporating}, can still overfit on \dataset. We therefore perform grid search on the En$\to$De validation set in \dataset\ and obtain a \textit{Tiny} configuration that works surprisingly well.
%For details of these model sizes and training settings, see Appendix~\ref{app:size}.
%A \textit{Tiny} model has 4 encoder layers and 4 decoder layers. The dimensions of input layer, output layer, and inner feed-forward layer are set to 128, 128, and 256, respectively. The number of attention heads is set to 4.
%

We use Adam with $\beta_1=0.9$, $\beta_2=0.98$ for model optimization. We start training with a warm-up phase (2,000 steps) where we linearly increase the learning rate from $10^{-7}$ to 0.005. Thereafter we decay the learning rate proportional to the number of updates. Each training batch contains at most 4,096 source/target tokens. We set label smoothing weight to 0.1, dropout to 0.3. We follow \cite{zhang2020neural} to early-stop the training if validation loss does not improve for ten epochs. We average the last ten checkpoints for inference as in \cite{vaswani2017attention} and \cite{wu2018pay}. We perform beam search with beam size set to 5. We report 4-gram BLEU and METEOR scores for all test sets. 
%\textmd{Multi-bleu.perl} was used to compute 4-gram BLEU scores for all test sets. 
All models are trained and evaluated on one single machine with two Titan P100 GPUs. 

% We use multi-bleu.perl\footnote{https://github.com/moses-smt/mosesdecoder/tree/RELEASE-4.0/scripts/ generic/multi-bleu.perl} to compute 4-gram BLEU scores for all test sets.

\subsection{Baselines}

Our baselines can be categorized into three types:
\begin{itemize}[wide=0\parindent,noitemsep]
    \item The text-only \textbf{Transformer};
    \item The conventional MMT models: \textbf{Doubly-ATT} and \textbf{Imagination};
    \item The retrieval-based MMT models: \textbf{UVR-NMT}.
\end{itemize}
Details of these methods can be found in Section~\ref{sec:related}.
For fairness, all the baselines are implemented by ourselves based on FairSeq~\cite{ott2019fairseq}. 
We use top-5 retrieved images for both UVR-NMT and our \model.
We also consider two more recent state-of-the-art conventional methods for reference: \textbf{GMNMT}~\citep{yin2020novel} and \textbf{DCCN}~\citep{lin2020dynamic}, whose results are reported as in their papers. 
Note that most MMT methods are difficult (or even impossible) to interpret. %(e.g., Doubly-ATT, Imagination, and DCCN).
While there exist some interpretable methods (e.g., UVR-NMT) that contain gated fusion layers similar to ours, they perform sophisticated transformations on visual representation before fusion,
which lowers the interpretability of the gating matrix.
For example, in the gated fusion layer of UVR-NMT, we observe that the visual vector is order-of-magnitude smaller than the textual vector. As a result, interpreting gating weight is meaningless because visual vector has negligible influence on the fused vector. 

\subsection{Results}

\begin{table*}[]
	\centering
	\resizebox{\textwidth}{!}{
		\begin{tabular}{c|l|c|ccc|c|ccc}
			\toprule
			\multirow{2}{*}{\#} &\multirow{2}{*}{\textbf{Model}} & \multicolumn{4}{c|}{\textbf{En$\to$De}} & \multicolumn{4}{c}{\textbf{En$\to$Fr}} \\ 
			& & \textbf{\#Params} & \textbf{Test2016} & \textbf{Test2017} & \textbf{MSCOCO} &  \textbf{\#Params} &\textbf{Test2016} & \textbf{Test2017} & \textbf{MSCOCO} \\ 
			\midrule
			\multicolumn{10}{c}{\textit{Text-only Transformer}} \\ \hline
			1 &\textbf{Transformer}-Base & 49.1M & 38.33 & 31.36 & 27.54 & 49.0M & 60.60 & 53.16 & 42.83 \\
			2 &\textbf{Transformer}-Small & 36.5M & 39.68 & 32.99 & 28.50 & 36.4M & 61.31 & 53.85 & 44.03 \\
			3 &\textbf{Transformer}-Tiny & 2.6M  & 41.02& 33.36& 29.88 & 2.6M &61.80 & 53.46 & 44.52 \\ \hline
			\multicolumn{10}{c}{\textit{Existing MMT Systems}} \\ \hline
			4 &\textbf{GMNMT}$^\spadesuit$ & 4.0M & 39.8  & 32.2 & 28.7 & - & 60.9 & 53.9 & - \\
			5 &\textbf{DCCN}$^\spadesuit$ & 17.1M & 39.7  & 31.0 & 26.7 & 16.9M & 61.2  & 54.3 & \textbf{45.4} \\ 
			6 &\textbf{Doubly-ATT}$^\spadesuit$ & 3.2M & 41.45 & \textbf{33.95} & 29.63 & 3.2M & 61.99 & 53.72 & 45.16\\
			7 &\textbf{Imagination}$^\spadesuit$ & 7.0M & 41.31 & 32.89 & 29.90 & 6.9M & 61.90 & 54.07 & 44.81 \\ 
			8 &\textbf{UVR-NMT}$^\diamondsuit$ & 2.9M & 40.79 & 32.16 & 29.02 & 2.9M & 61.00 & 53.20 & 43.71 \\ \hline
			\multicolumn{10}{c}{\textit{Our MMT Systems}} \\ \hline
			%9 & \textbf{Balanced Fusion}$^\spadesuit$ & 2.8M & 39.48 & 30.39 & 27.51 & 2.8M & 57.36 & 49.86 & 42.03 \\
			9 &\textbf{Gated Fusion}$^\spadesuit$ & 2.9M & \textbf{41.96} & 33.59 & 29.04  & 2.8M & 61.69 & \textbf{54.85} & 44.86 \\
			10 &\textbf{\model}$^\diamondsuit$ & 2.9M & 41.45 & 32.94 & \textbf{30.01} & 2.9M & \textbf{62.12} & 54.39 & 44.52 \\
		%	11 &\textbf{\model-uvr}$^\diamondsuit$ & 3.3M & 40.61 & 32.77 & 29.30 & 3.2M & 61.17 & 54.18 & 45.07 \\
			\bottomrule
		\end{tabular}
	}
	\caption{BLEU scores on \dataset. Results in row 4 and 5 are taken from the original papers. $\spadesuit$ indicates conventional MMT models, while $\diamondsuit$ refer to retrieval-based models. Without further specified, all our implementations are based on the \textit{Tiny} configuration. }
	\label{tab:multi30k}
\end{table*}
		%$\ddagger$/$\dagger$: significantly better than Transformer-Tiny ($p<$ 0.01/0.05), **/*: significantly better than Imagination ($p<$ 0.01/0.05)}

Table~\ref{tab:multi30k} shows the BLEU scores of these methods on the \dataset\ dataset.
%We first observe that the widely-used MMT baseline --- Transformer-Base is a weak baseline that overfits the \dataset\  dataset. 
From the table, we see that
although we can replicate similar BLEU scores of Transformer-Base
as reported in \cite{gronroos2018memad, ive-etal-2019-distilling},
these scores (Row 1) are significantly outperformed by
Transformer-Small and Transformer-Tiny, which have fewer parameters.
%than Transformer-Base.
%This shows that large models like Transformer-\textit{Base} lead to overfitting on \dataset
This shows that
%large models like
Transformer-Base could overfit the \dataset\ dataset. %whose parameters are about $1/20$ in number of Transformer-\textit{Base}, 
Transformer-Tiny, whose number of parameters is about $20$ times smaller than that of Transformer-Base,
is more robust and efficient in our test cases. We therefore use it as the base model for all our MMT systems in the following discussion.

% 
%On the basis of Transformer-tiny, our implementation of Doubly-ATT, Imagination, and UVR-NMT significantly outperform the numbers achieved based on Transformer-base~\cite{helcl2018doubly,arslan2018doubly,zhang2020neural}. 
Based on the Transformer-tiny model, both our proposed models (Gated Fusion and RMMT) and baseline MMT models (Doubly-ATT, Imagination and UVR-NMT)  significantly outperform the state-of-the-arts (GMNMT and DCCN) on En$\to$De translation.
%Also they achieve close and on-par with them on En$\to$Fr translation.  
%However, none of these methods can consistently achieve better performance than the ever-strong Transformer-Tiny baseline (row 3).
However, 
the improvement of all these methods (Rows 4-10) over the base Transformer-Tiny model (Row 3) is very marginal.
This shows that 
visual context might not be as important as we expected 
for translation, at least on datasets we explored.
% Note that such \textit{modest} gains after introducing multimodal context have been widely reported, for both LSTM-based models~\cite{mmt2016,mmt2017} and Transformer-based models~\cite{mmt2018,gronroos2018memad}, and in terms of both automatic metrics (e.g., BLEU and METEOR) and human evaluation~\cite{mmt2018}. 

% We further conduct experiments on the VaTex dataset and evaluate all the models w.r.t. the METEOR scores (see Appendix~\ref{app:vatex} and~\ref{app:meteor}). 

We further evaluate all the methods on the METEOR scores (see Appendix~\ref{app:meteor}).
We also run experiments on the VaTex dataset 
(see Appendix~\ref{app:vatex}). 
Similar results are observed as Table~\ref{tab:multi30k}.
%All the results are given in .
Although various MMT systems have been proposed recently, a well-tuned model that uses text only remain competitive. 
This motivates us to revisit the importance of visual context for translation in MMT models.

%% file: 5-analysis.tex
\section{Model Analysis}
%Inspecting gains reported in the last section closer, 
Taking a closer look at the results given in the previous section, we are surprised by the observation
that our models learn to \textit{ignore} visual context when translating (Sec~\ref{sec:probe}). This motivates us to revisit the contribution of visual context in MMT systems (Sec~\ref{sec:revisit}).
Our adversarial evaluation shows that adding model regularization achieves comparable results as incorporating visual context. 
%We devise adversarial experiments that replacing visual features with random noise and observe similar gains. Stimulating us to hypothesize that visual context actually serves as some kind of regularization. To further test our hypothesis, in a more radical setup, we deprive all visual context from a MMT system, and augment it with weight decay. We also observe compatible improvements. 
Finally, we discuss when visual context is needed (Sec~\ref{sec:limited}) and how these findings could benefit future research.

%We then design adversarial experiments to probe the need of visual context in MMT . 

%Our results complement previous findings and suggest that under limited textual context, MMT models are capable of leveraging the visual input to generate better translations (Section~\ref{sec:revisit})\zy{basically findings of Ozan et al., 2019, highlight how we complement their findings}. 
%However, it remains unclear how visual context improve translation when textual context is already sufficient enough for translating. We devise experiments to verify our hypothesis that under sufficient textual context, visual context help translation in a way similar to regularization functions (Section~\ref{label}). 

\subsection{Probe the need for visual context in MMT}
\label{sec:probe}

\begin{table}[]
	\centering
	\small
	\begin{tabular}{l | c c }
		\toprule
		Multi30k & Gated Fusion & \model   \\
		\midrule
		En$\to$De \\ 
		$\quad $ Test2016 & 4.5e-21  & 8.6e-13\\
		$\quad $ Test2017 & 7.0e-17  & 4.0e-13 \\
		$\quad $ MSCOCO & 9.7e-21  & 3.5e-14 \\ \hline
		En$\to$Fr \\ 
		$\quad $ Test2016 & 1.6e-18 & 1.1e-11\\
		$\quad $ Test2017 & 7.2e-15 & 5.0e-12\\
		$\quad $ MSCOCO & 2.3e-18  & 5.3e-13\\ 
		%\multicolumn{3}{l}{VaTex En$\to$Zh} \\
		%$\quad $ Test & $25.0\%$  & $33.89\%$
		\bottomrule
	\end{tabular}
	\caption{Micro-averaged gating weight $\overline{\Lambda}$ on \dataset.}
	\label{tab:awareness}

\end{table}

To explore the need for visual context in our models, we focus on the interpretable component: the gated fusion layer (see Equation~\ref{eq:fusion} and \ref{eq:rmmtfusion}). Intuitively, a larger gating weight $\Lambda_{ij}$ indicates the model learns to depend more on visual context to perform better translation. 
We quantify the degree to which visual context is used by the micro-averaged gating weight $\overline{\Lambda} = \sum_{m=1}^M \operatorname{sum}(\Lambda^m) / (d \times V)$.
Here $M$, $V$ are the total number of sentences and words in the corpus, respectively. $\operatorname{sum}(\cdot)$ add up all elements in a given matrix, and $\overline{\Lambda}$ is a scalar value ranges from 0 to 1. A larger $\overline{\Lambda}$ implies more usage of the visual context.
%However,  $\overline{\lambda}$ does not consider the original representation, we propose an extension called \textit{visual awareness} (denoted as $\text{Awareness} \in [0, \infty]^{T \times d}$). $\text{Awareness}$ is defined as $abs(\mathbf{H}_{\text{image}} )/ abs(\mathbf{H}$, where $abs()$ is the absolute value function. Visual awareness measures how much visual representation will affect the final fused representation. A large $\text{Awareness}$ indicates a high dependency on visual information for translation.
%\textbf{in which each $\text{Awareness}_i  \in [0, \infty]^d$}

We first study models' behavior after convergence. 
From Table~\ref{tab:awareness}, we observe that $\overline{\Lambda}$ is negligibly small, suggesting that both models learn to discard visual context.
%~\footnote{Note that in \model-uvr,  visual context is discarded during sophisticated transformation layer $f_{\theta}(*)$ (i.e., before fusion layer), thus we omit the discussion of its $\overline{\Lambda}$}.
%In other words, visual context does not provide additional useful information for translation, or at least the offering is much less than what was previously thought.
In other words, visual context may not be as important for translation as previously thought.
Since $\overline{\Lambda}$ is insensitive to outliers (e.g., large gating weight at few dimensions),
we further compute $p(\Lambda_{ij} > 1e$-$10)$: percentage of gating weight entries in $\Lambda$ that are larger than $1e$-$10$. With no surprise, we find that on all test splits $p(\Lambda_{ij} > 1e$-$10)$ are always zero, which again shows that visual input is not used by the model in inference.
%Worse yet, we suspect that $\overline{\lambda}$ are in fact zero, such negligibly small values are results of numerical precision issue. 
%Worse yet, non-zero gating values are generally within $[10^{-10}, 10^{-40}]$. \bi{this result is not convincing enough. actually u need to compute the percentage of the value from the vision part over the norm/abs of each dim. because it is possible that the scale of the two vectors are very different.} Such negligibly small values indicate that even though the visual context is exploited by the model, it is orders-of-magnitude less important than previously thought. 

% \begin{figure}
%     \centering
%     \includegraphics[width=\linewidth]{figures/fig-ende.pdf}
%     \caption{Training dynamic analysis on Test2016 of Multi30k En-de translation, starting from Epoch 1.}
%     \label{fig:dynamic:bleu}
% \end{figure}

\begin{figure}[]

	\centering
	\subfigure[En$\to$De.]{\includegraphics[width=0.8\linewidth]{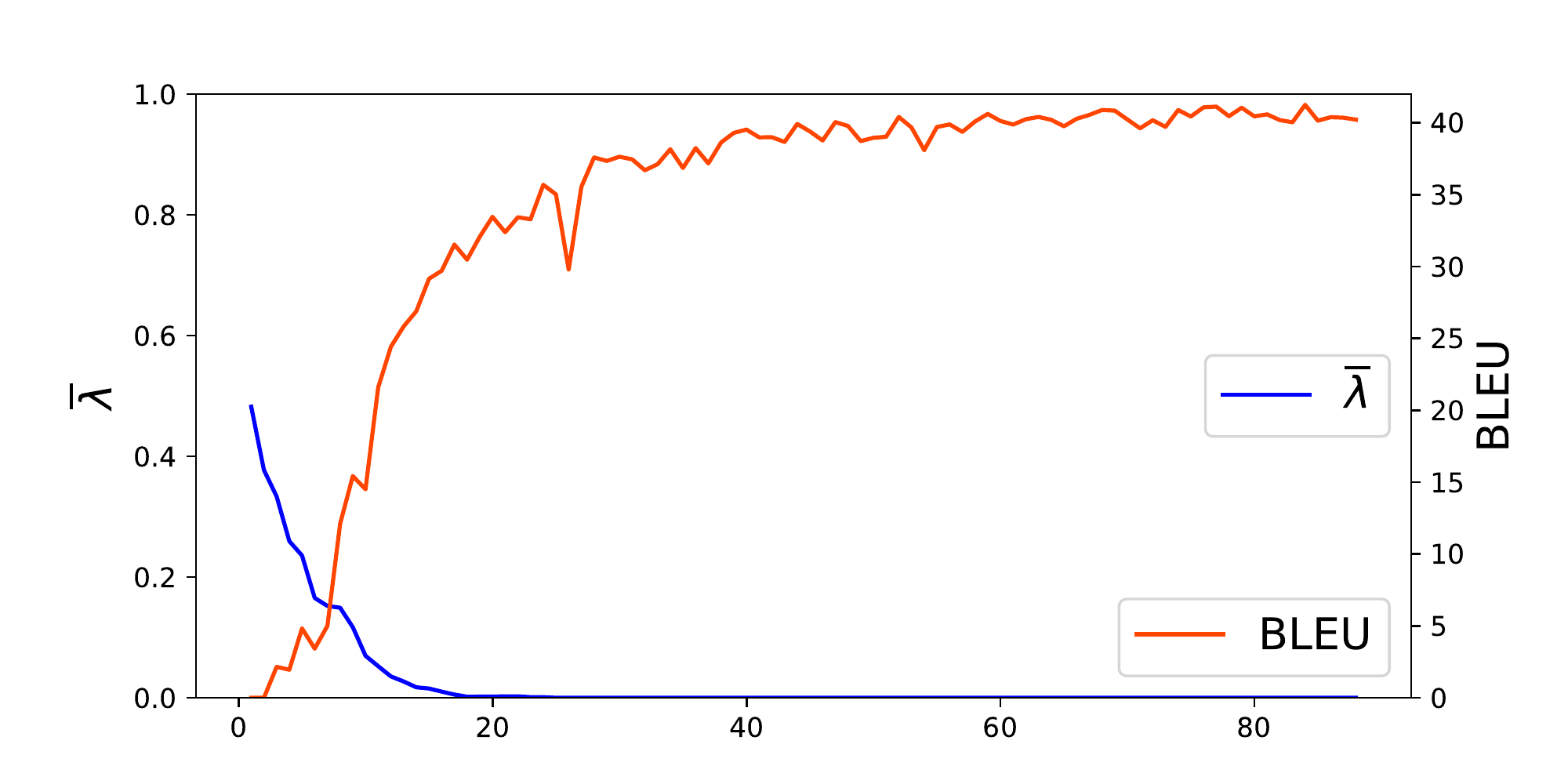}}
	\vspace{-0.5\baselineskip}
	
	\subfigure[En$\to$Fr.]{\includegraphics[width=0.8\linewidth]{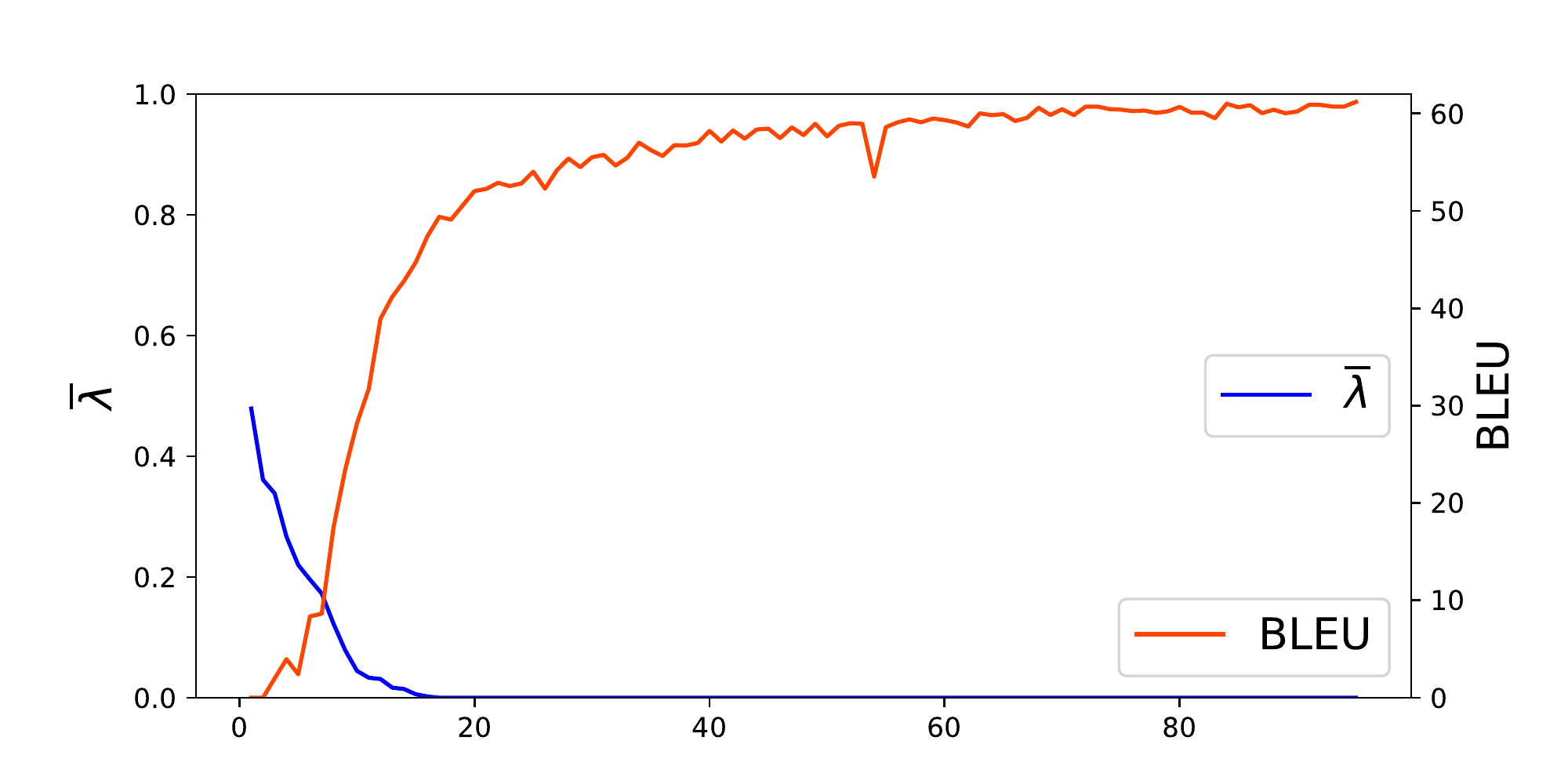}}
	
	\caption{Training dynamic of Multi30k En$\to$de and En$\to$Fr translation, from Epoch 1. }
	\label{fig:dynamic}
\end{figure}
%Inspecting closer, 
The Gated Fusion's training process also shed some light on how the model accommodates the visual information during training. Figure~\ref{fig:dynamic} (a) and (b) shows how $\overline{\Lambda}$ changes during training, from the first epoch.
We find that, Gated Fusion starts with a relatively high $\overline{\Lambda}$ ($>$0.5), but quickly decreases to $\approx 0.48$ after the first epoch. As the training continues, $\overline{\Lambda}$ gradually decreases to roughly zero. 
% The learning curve reveals how visual context is used in Gated Fusion. 
In the early stages, the model relies heavily on images, %because visual features consist of structured signals extracted from pre-trained ResNet-50, whereas the textual encoder is randomly initialized and barely carry any information about source text
possibly because they could provide meaningful features extracted from a pre-trained ResNet-50 CNN, while the textual encoder is randomly initialized.
Compared with text-only NMT, utilizing visual features 
lowers MMT models' trust in the hidden representations generated from the textual encoders. 
As the training continues, the textual encoder learns to represent source text better and the importance of visual context gradually decreases. 
In the end, the textual encoder carries sufficient context for translation and supersedes the contributions from the visual features.
Nevertheless, this doesn't explain the superior performance of the multimodal systems (Table~\ref{tab:multi30k}). We speculate that visual context is acting as regularization that helps model training in the early stages. We further explore this hypothesis in the next section.

%Recall the whole learning process, we find that visual context is only helpful during training and will be discarded after model converges.
%They are acting like some kind of regularization methods that regularize model's training in early stage. We further explore this hypothesis in the next section \li{remove the whole paragraph. Too redundant}.
%learns to translate without ``seeing'' images. Because with such a magnitude smaller weight, most (if not all) visual knowledge is discarded during fusion. 
%However, although negligibly small, $\overline{\lambda}$ is never precisely zero (i.e. the model will not discard visual information completely).
%This motivates us to revisit the role of visual information in MMT. 
%We conjecture that the visual information is functioning like some structured noise that slightly alters the model's hidden representation. 
%We observe a similar trend in RMMT that the retriever's recall quickly decrease to near zero over time. 

\subsection{Revisit need for visual context in MMT}
\label{sec:revisit}

\begin{table*}[]
	\centering
	\resizebox{\textwidth}{!}{
		\begin{tabular}{c|l|ccc|ccc}
			\toprule
			\multirow{2}{*}{\#} &\multirow{2}{*}{\textbf{Model}} & \multicolumn{3}{c|}{\textbf{En$\to$De}} & \multicolumn{3}{c}{\textbf{En$\to$Fr}} \\ 
			& &\textbf{Test2016} & \textbf{Test2017} & \textbf{MSCOCO} &\textbf{Test2016} & \textbf{Test2017} & \textbf{MSCOCO} \\ 
			\toprule
			1 &\textbf{Transformer} & 41.02 & 33.36 & 29.88 & 61.80 & 53.46 & 44.52 \\ \hline
			2 &\textbf{Doubly-ATT} & 41.53(+0.08) & 33.90(-0.05) & 29.76(+0.15) & 61.85(-0.35) & 54.61(+0.46) & 44.85(-0.80) \\
			3 &\textbf{Imagination} & 41.20(-0.11) & 33.32(+0.42) & 29.92(+0.02) & 61.28(-0.62) & 53.74(-0.33) & 44.89(+0.08)\\
			4 &\textbf{Gated Fusion} & 41.53(-0.45) & 33.52(-0.07) & 29.87(+0.83) & 61.58(-0.11) & 54.21(-0.64) & 44.88(+0.02) \\
			% 5 &\model-Static & 41.60 & 32.98 & 29.54 &  &  &  \\
			\bottomrule
		\end{tabular}
	}
	\caption{BLEU scores on \dataset\ with randomly initialized visual representation. Numbers in parentheses indicate the relative improvement/deterioration compared with the same model with ResNet feature initialization.}
	\label{tab:random}
\end{table*}

%The phenomena of ``zero-valued gating vector'' in Gated Fusion and \model\ suggests that visual context does not offer additional useful information in MMT, or at least the offering is much less than what was previously thought. 
%Given the observation from Section~\ref{sec:probe}, we hypothesize that the gains of MMT systems are due to some regularization effects, where the additional information is treated as random noise that lowers the model's trust in the hidden representations generated from the neural encoders, resulting in smaller network weights and a more robust network that has lower generalization error~\cite{an1996effects}

%that both previous papers and our experiments observe similar gains of MMT systems when they are compared with text-only baselines, we want to investigate where these gains come from. Our hypothesis is that the improvements are due to some regularization effects, where the additional information is treated as random noise that lowers the model's trust in the hidden representations generated from the neural encoders, resulting in smaller network weights and a more robust network that has lower generalization error.
In the previous section, we hypothesize that the gains of MMT systems come from some regularization effects.
To verify our hypothesis, 
we conduct experiments based on two widely used regularization techniques: random noise injection~\citep{bishop1995training} and weight decay~\citep{hanson1989comparing}. The former simulates the effects of assumably uninformative visual representations and the later is a more principled way of regularization that does not get enough attention in the current hyperparameter tuning stage. Inspecting the results, we find that applying these regularization techniques achieves similar gains over the text-only baseline as incorporating multimodal information does.
% Next we study how much these regularization techniques can improve the text-only baseline.

For random noise injection,
we keep all hyper-parameters unchanged but replace visual features extracted using ResNet with randomly initialized vectors, which are noise drawn from a standard Gaussian distribution. A MMT model equipped with ResNet features is denoted as a \textit{ResNet-based model}, while the same model with random initialization is denoted as a \textit{noise-based model}. 
%In other words, the noise was added once per training sequence as in \cite{graves2013speech} rather than at every timestep. 
We run each experiment three times and report the averaged results. 
Note that values in parentheses indicate the performance gap between the ResNet-based model and its noise-based adversary.

Table~\ref{tab:random} shows BLEU scores on the \dataset\ dataset.
Each column in the table corresponds to a test set ``contest''.
% From the table, we observe that
% noise-based models achieve similar gains over the Transformer baseline as ResNet-based models.
From the table, we observe that, among 18 (3 methods $\times$ 3 test sets $\times$ 2 tasks) contests with the Transformer model (row 1), noise-based models (rows 2-4) achieve better performance 13 times, while ResNet-based models win 14 cases.
This shows that noise-based models perform comparably with ResNet-based models.
A further comparison between noise-based models and ResNet-based models shows that they are compatible after 18 contests, in which the former wins 8 times and the latter wins 10 times. 
%tie in 18 contests, each wins 9 times.

We observe similar results when repeating above evaluation using METEOR (Tabel~\ref{tab:meteor_random} %in Appendix~\ref{app:meteor}
) and on VaTex (Table~\ref{tab:vatex} %in Appendix~\ref{app:vatex}
).
These observations deduce that random noise could function as visual context. 
%Injecting noise into a neural network makes it difficult for the network to fit individual data points precisely, and hence it reduces overfitting~\citep{bishop1995neural}
%\li{remove the sentence}.
%In terms of MMT, added noise or visual context makes the translations of sentences in \dataset, which are short and repetitive~\citep{caglayan2019probing}, more challenging.
In MMT systems, adding random noise or visual context can help
reduce overfitting~\citep{bishop1995neural} when translating sentences in \dataset, which are short and repetitive~\citep{caglayan2019probing}.
Moreover, we find that %the $\ell_2$ norm of model weights in Gated Fusion with ResNet features (with random noise) is only 97.7\%(95.2\%) of that in Transformer on En$\to$De
the $\ell_2$ norm of model weights in ResNet-based Gated Fusion and noise-based Gated Fusion are only 97.7\% and 95.2\% of that in Transformer on En$\to$De, respectively. This further %substantiates 
verifies our speculation that, as random noise injection~\cite{an1996effects}, %visual context can help model smooth weights and generalize better
visual context can help weight smoothing and improve model generalization.% does.%, just like adding random noise does~\cite{an1996effects}. 

Further, we regularize the models with weight decay. %We further\lpk{is it further? means on top of random noise or is it separate} regularize models with weight decay. 
%Weight decay has not been widely adopted in NMT because NMT datasets generally contain millions of sentences and are less likely to overfit. Even in low-resource translation~\citep{wu2018pay}, researchers are very cautious with the usage of weight decay (e.g., 0.0001). 
We consider three models: the text-only Transformer, the representative existing MMT method Doubly-ATT, and our Gated Fusion method. 
Figure~\ref{fig:weight_decay_bleu} and ~\ref{fig:weight_decay_meteor} (in Appendix~\ref{app:meteor}) show the BLEU and METEOR scores of these methods on En→De translation as weight decay rate changes, respectively. 
We see that 
the best results of 
the text-only Transformer model with fine-tuned weight decay are comparable or even better than that of the MMT models Doubly-ATT and Gated Fusion that utilize visual context.
This again shows that
visual context is not as useful as we expected
and it essentially plays the role of regularization.

% We see that 
% with a large penalty on model weights (0.1), the text-only Transformer (\textcolor{RED}{\textbf{red}} line with triangle) can also exhibit performance that is comparable to or even superior than MMT models (either with or without weight decay).
%Dounly-ATT (\textcolor{GREEN}{\textbf{green}} line with $\blacksquare$) and Gated Fusion (\textcolor{BLUE}{\textbf{blue}} line with $\bigstar$). 
% On the other hand, introducing more regularization into MMT models always deteriorates the performance. 
% One plausible explanation is that the visual context already serves as regularization in MMT, thus imposing more regularization will lead to under-fitting. 
%Together with random noise experiments, the results of adding weight decay further support our hypothesis that multimodal context serves as a kind of regularization. 

\begin{figure*}[]
    \centering
    \begin{minipage}{0.33\linewidth}
    \subfigure[Test2016.]{
    \includegraphics[width=\linewidth]{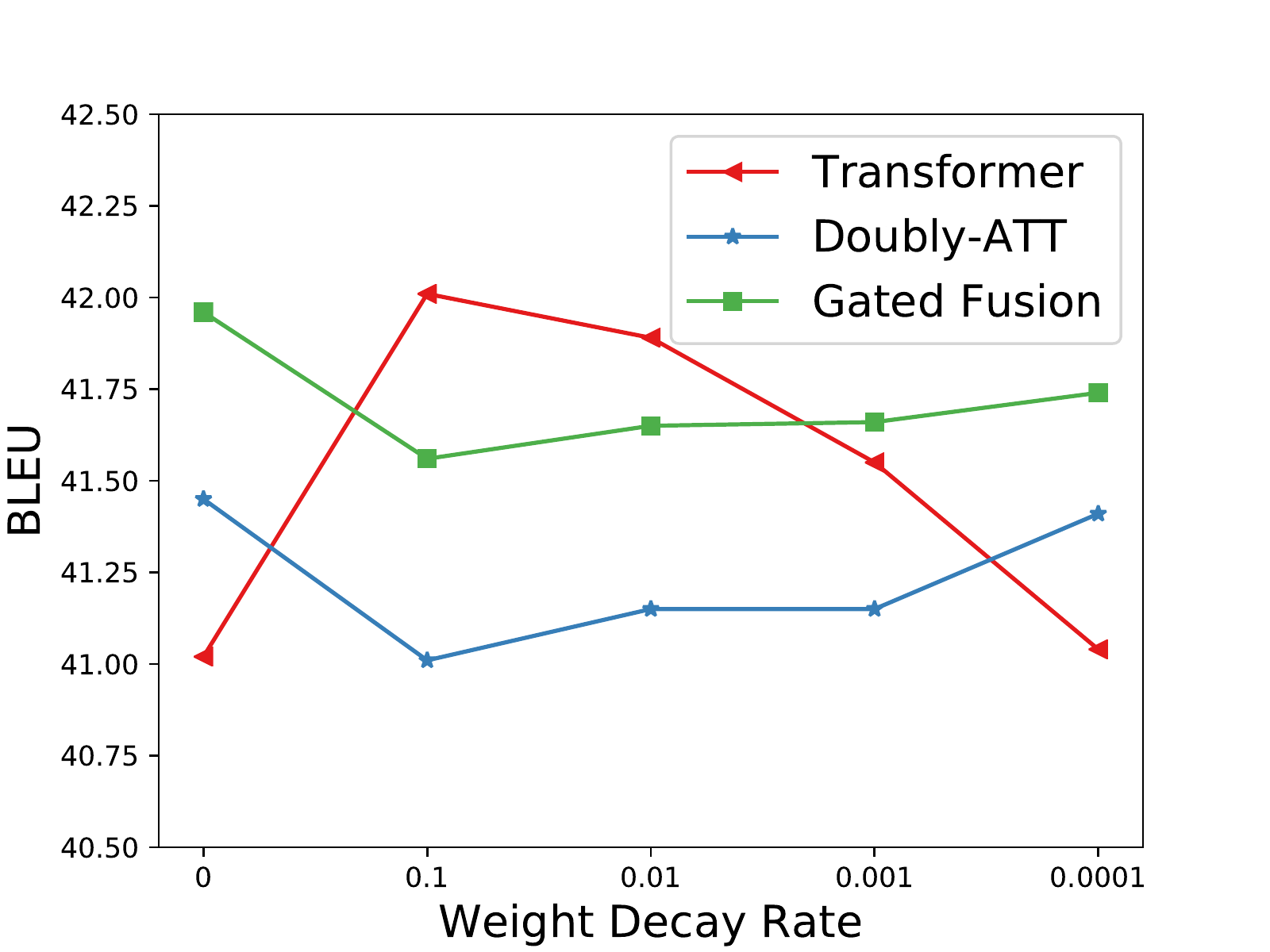}
    }
    \end{minipage}%
    \begin{minipage}{0.33\linewidth}
    \subfigure[Test2017.]{
    \includegraphics[width=\linewidth]{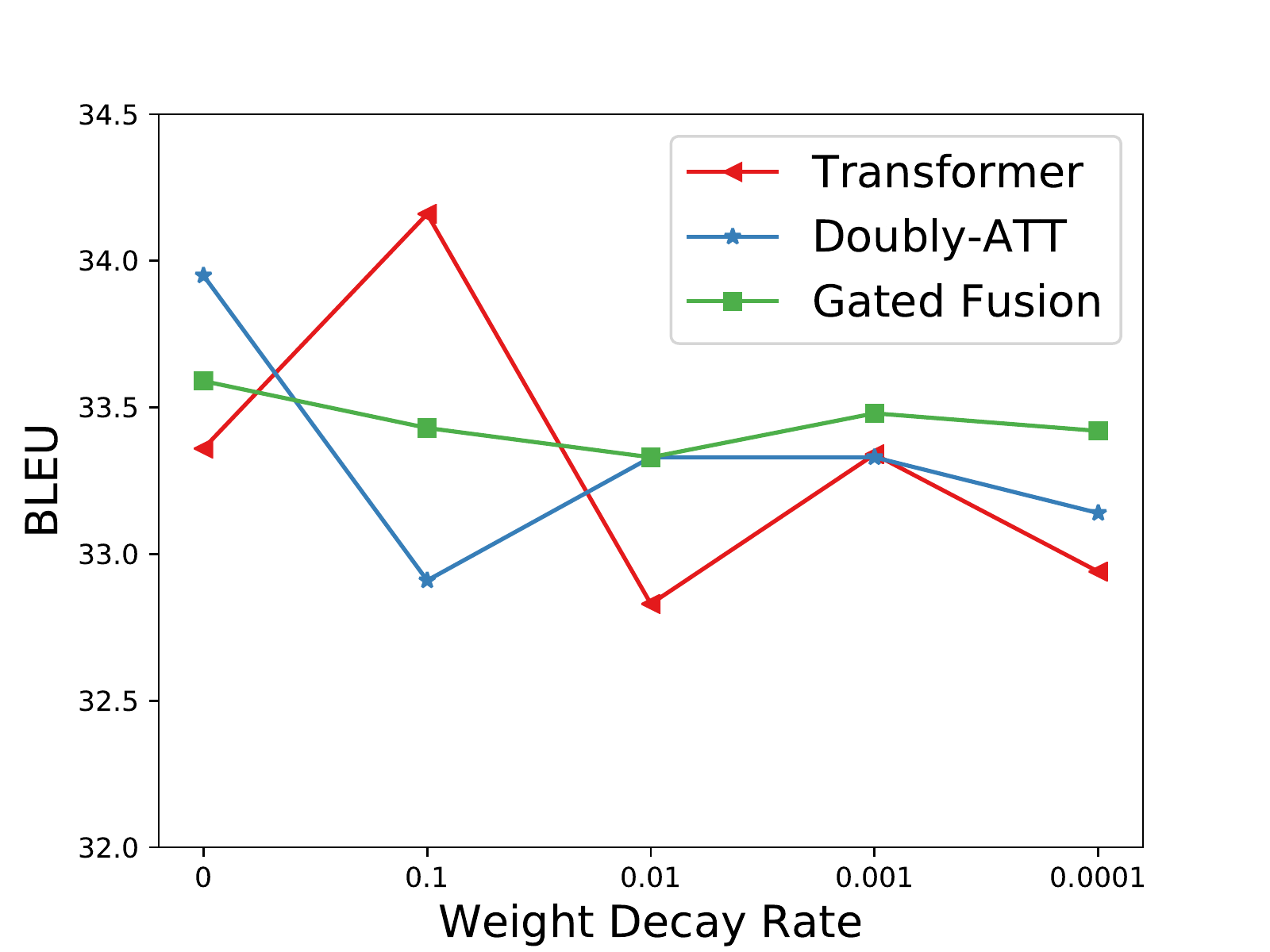}
    }
    \end{minipage}%
    \begin{minipage}{0.33\linewidth}
    \subfigure[MSCOCO.]{
    \includegraphics[width=\linewidth]{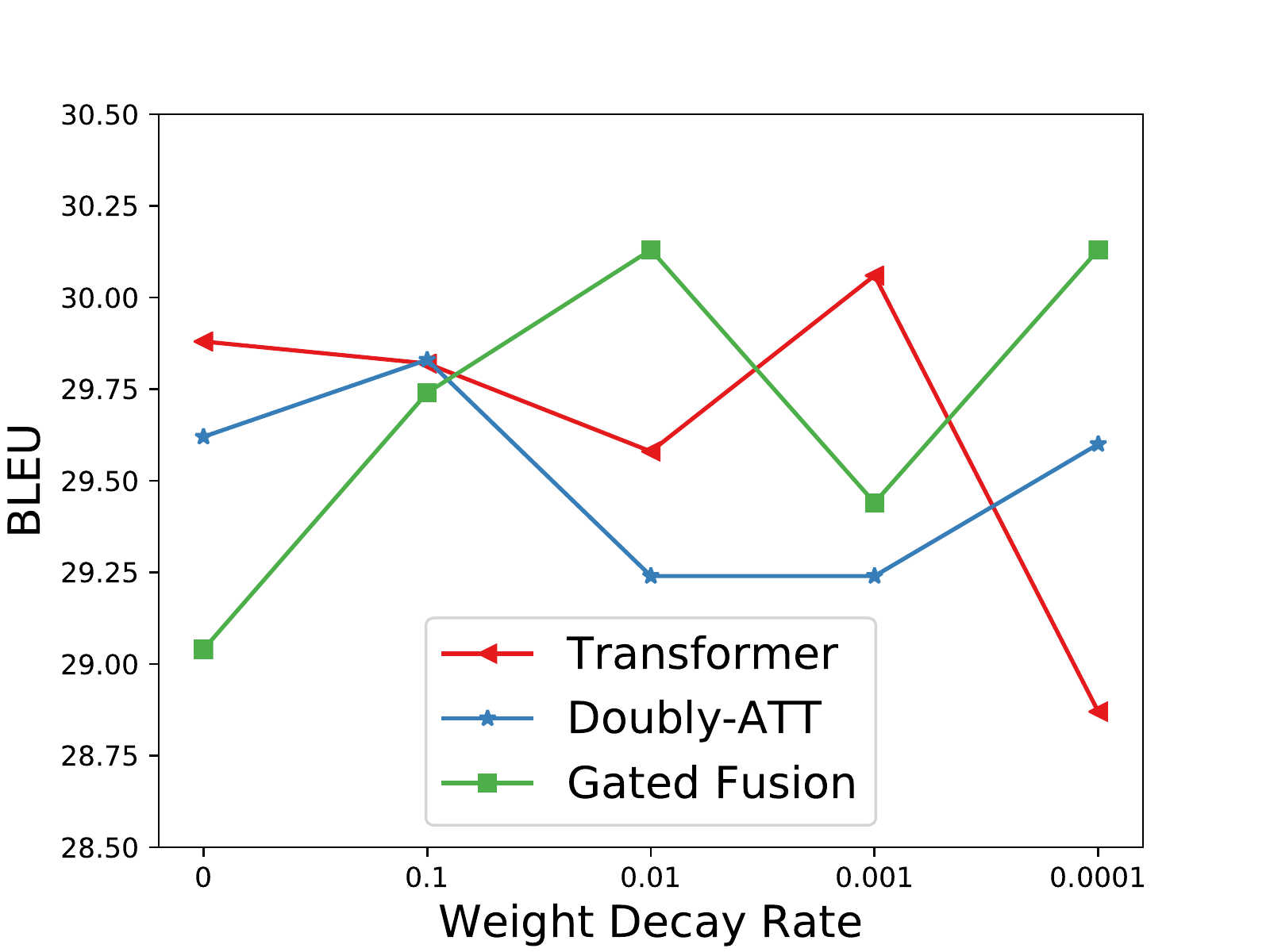}
    }
    \end{minipage}
    \caption{BLEU score curves on En$\to$De translation with different weight decay rate.}
    \label{fig:weight_decay_bleu}
\end{figure*}

% Our findings above indicate that visual context does not offer additional useful information when translating, at least on datasets we explored. The major pitfall, however, is current benchmarks being too simple, such that source sentences can provide sufficient context for translation. In the next section, we discuss when visual context is needed in MMT and how that would guide future research.

\subsection{When is visual context needed in MMT}
\label{sec:limited}

\begin{table}[]
	\centering
	\small
	\resizebox{\linewidth}{!}{
	\begin{tabular}{c | l c c c }
		\toprule
		& &  			BLEU &  METEOR  & $\overline{\Lambda}$\\
		\midrule
		%\multicolumn{2}{l}{Test2016} \\ 
		1 & Transformer & 11.39 & 35.53 & -  \\
		2 & $\quad $ +weight decay 0.1 & 11.66 & 35.95 & - \\
		\midrule
		& \multicolumn{4}{l}{w. ResNet features}  \\
% 		\midrule
		3 & $\quad $ Gated Fusion& 14.79 & 40.41 & 0.047  \\
		4 & $\quad $ \model& 16.67 & 43.62  & 0.011 \\
		\midrule
		& \multicolumn{4}{l}{w. random noise}  \\
% 		\midrule
		5 & $\quad $ Gated Fusion& 11.40 & 35.44 & 0.032  \\
		6 & $\quad $ \model& 12.08 & 37.60 & 0.010 \\
		\bottomrule
	\end{tabular}
}
	\caption{Adversarial evaluation with limited textual context on \dataset\ En-De Test2016.}
	\label{tab:masked}
\end{table}

Despite the less importance of visual information we showed in previous sections,
there also exist works that support its usefulness. For example,
\citet{caglayan2019probing} experimentally show that, with limited textual context
(e.g., masking some input tokens), MMT models will utilize the visual input for translation. 
This further motivates us to investigate when visual context is needed in MMT models.
%In this section, w
We conduct experiment with a new masking strategy that does not need any entity linking annotations as in \citet{caglayan2019probing}.
Specifically, we follow \citet{tan2020vokenization} to collect a list of \textit{visually grounded tokens}. 
A visually grounded token is the one that has more than 30 occurrences in the \dataset\ dataset with stop words removed. Masking all visually grounded tokens will affect around 45\% of tokens in \dataset. 

\begin{table*}[hbt!]
	\centering
	\resizebox*{\linewidth}{!}{
		\begin{tabular}{c c l l }
			\toprule
			\begin{minipage}{0.03\linewidth}
				(a) 
			\end{minipage}
			\begin{minipage}{.2\linewidth}
				\includegraphics[width=\linewidth]{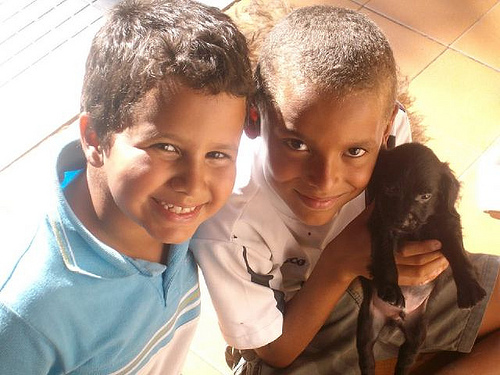}
			\end{minipage}
			&
			\begin{minipage}{0.05\linewidth}
				SRC:   \\ NMT: \\ \\ MMT: \\ \\ REF: \\
			\end{minipage}
			&
			\begin{minipage}{0.7\linewidth}
				two \underline{young boys pose} with a puppy for a \underline{family picture}\\
				zwei \sout{braune hunde spielen} mit einem \sout{spielzeug} für einen \sout{tennisball}  \\
				\textit{(two brown dogs play with a toy for a tennis ball)} \\
				zwei \textbf{kleine jungen posieren} mit einem \textbf{welpen} für \sout{ein foto} \\
				\textit{(two little boys pose with a puppy for a photo)} \\
				zwei kleine jungen posieren mit einem welpen für eine familienfoto \\
				\textit{(two little boys pose with a puppy for a family photo)}
			\end{minipage}
			\\
			\midrule
			\begin{minipage}{0.03\linewidth}
				(b) 
			\end{minipage}
			\begin{minipage}{.2\linewidth}
				\includegraphics[width=\linewidth]{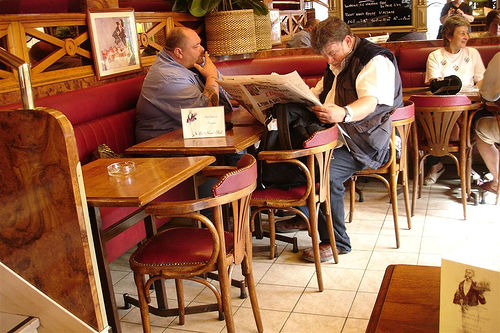}
			\end{minipage}
			&
			\begin{minipage}{0.05\linewidth}
				SRC:   \\ NMT: \\ \\ MMT: \\ \\ REF: \\
			\end{minipage}
			&
			\begin{minipage}{0.7\linewidth}
				\underline{	two men sitting} in a \underline{restaurant} \\
				zwei \sout{kinder spielen} in einem \sout{springbrunnen}  \\
				\textit{(two children are playing in a fountain)} \\
				zwei \sout{frauen} sitzen in einem \textbf{restaurant} \\
				\textit{(two women are sitting in a restaurant)} \\
				zwei männer sitzen in einem restaurant \\
				\textit{(two men are sitting in a restaurant)}
			\end{minipage}
			\\
			\bottomrule
		\end{tabular}
	}
	\caption{Case studies under limited textual input. We use underline to denote masked tokens, and \sout{strikethrough} (\textbf{bold}) font to denote \sout{incorrect} (\textbf{correct}) lexical choices. We use Gated Fusion for analysis. }
	\label{table:cases}
\end{table*}

Table~\ref{tab:masked} shows the adversarial study with visually grounded tokens masked.
In particular, we select Transformer, Gated Fusion and \model\ as representative methods.
%Repeating our adversarial evaluation using weight decay (Table~\ref{tab:masked}, row 2) and random noise (row 5,6), however, we see that these regularization methods only marginally improve performance under limited textual input. 
From the table, we see that random noise injection (row 5,6) and weight decay (row 2) can only bring marginal improvement over the text-only Transformer model.
%On the other side, introducing meaningful visual inputs (ResNet features) can significantly improve translation, in terms of both BLEU and METEOR (see rows 1,3, and 4). 
However, ResNet-based models that utilize visual context significantly improve the translation results.
For example, \model\ achieves almost 50\% gain over the Transformer on the BLEU score.
%And both models' $\overline{\Lambda}$ ($0.01$-$0.5$) (rows 3 and 4) show clear distinctions from under sufficient textual context ($<$1e-10).
Moreover, both Gated Fusion and \model\ using ResNet features lead to a larger $\overline{\Lambda}$ value than that when textual context is sufficient as shown in Table~\ref{tab:awareness}.
Those results further suggest that visual context is needed when textual context is insufficient. 
%Note that besides input masking, sentences with incorrect, ambiguous, and gender-neutral words~\cite{frank2018assessing} can also be regard as under textual context, stressing the need for a more challenging benchmark to advance MMT research.
In addition to token masking, sentences with incorrect, ambiguous and gender-neutral words~\cite{frank2018assessing} might also need visual context to help translation.
%However, the current benchmark datasets (e.g., \dataset\ and VaTex) are rather simple, where source sentences could provide sufficient information for translation.
Therefore, to fully exert the power of MMT systems, we emphasize the need for a new MMT benchmark, in which visual context is deemed necessary to generate correct translation.% that requires more visual clues to solve for the community.

Interestingly, even with ResNet features, we observe a significant drop in both BLEU and METEOR scores compared with those in Table~\ref{tab:multi30k} and ~\ref{tab:meteor},
similar to that reported in~\citep{chowdhury2019understanding}.
The reason could be two-fold. On the one hand, there are many words that can not be visualized. For example,
in Table~\ref{table:cases} (a), although Gated Fusion can successfully identify the main objects in the image (``little boys pose with a puppy''), it fails to generate the more abstract concept ``family picture''.
On the other hand,
%correctly ``focusing on'' different image regions is challenging when translating different words.
when translating different words, it is difficult to capture correct regions in images.
For example, in Table~\ref{table:cases} (b), we see that Gated Fusion incorrectly generates the word \textit{frauen (women)} because it captures the woman at the top-right corner of the image.
% These results again show that visual input can only play an auxiliary role in MMT.

\subsection{Discussion}
%\bi{the insights are good but this para need to reorganize.}
Finally, we discuss how our findings might benefit future MMT research. 
First, %with the rapid development of the community, it's time to move away from the famous \dataset, and embrace a more challenging benchmark\li{remove with...}
a benchmark that requires more visual information than \dataset\ to solve is desired.
As shown in Section~\ref{sec:revisit}, sentences in \dataset\ are rather simple and %have precise meanings
easy-to-understand. Thus textual context could provide sufficient information for correct translation, % without the need of visual context, 
making visual modules relatively redundant in these systems. 
%making the introduction of visual context seems redundant, since the textual context is already sufficient to perform translation\li{remove}.
While the MSCOCO test set in \dataset\ contains ambiguous verbs and encourages models to use image sources for disambiguation, we still lack a corresponding training set.%\lpk{lack what data for training?}. 

Second, our methods can serve as a verification tool to investigate whether visual grounding is needed in translation for a new benchmark. %\lpk{we can say we will release our tool to the community as opensource if we are planing to do so.}

Third, %as shown in our case study (Table~\ref{table:cases} (b)), 
we find that visual feature selection is also critical for MMT's performance.
%However, obtain annotations for the feature selection process can be costly. Thus 
While most methods employ the attention mechanism to learn to attend relevant regions in an image, the shortage of annotated data could impair the attention module (see Table~\ref{table:cases} (b)). %\lpk{which experiments or sections back this claim?}
%based merely on the translation loss.%but do not provide corresponding supervision to guide the attention. 
Some recent efforts~\cite{yin2020novel,lin2020dynamic,caglayan-etal-2020-simultaneous} address the issue by feeding models with pre-extracted visual objects instead of the whole image.
However, these methods are easily affected by the quality of the extracted objects.
Therefore, a more effective end-to-end visual feature selection technique is needed,
which can be further integrated into MMT systems to improve performance.

%which is rarely considered in previous works.
%Most existing models use the attention mechanism to capture relevant regions in an image based on the translation loss (i.e., there is no direct supervision for feature selection) \li{remove sentence in ()}. 
%There are some recent efforts~\cite{yin2020novel,lin2020dynamic,caglayan2020simultaneous} that pre-extract
%regional features from images
%and show some promise.
%Moreover, there also exist works that try to introduce auxiliary distant visual supervision to help achieve grounding~\cite{tan2020vokenization}, which might be referenced to improve MMT.

%% file: 9-appendix.tex
\clearpage
\appendix
\section{Training Settings}
Table~\ref{tab:model_sizes} shows the configuration of different model sizes. 
\label{app:size}
\begin{table}[h!]
    \centering
    \resizebox{\linewidth}{!}{
        \begin{tabular}{l|c|c|c}
    	\toprule
    	Model component  & Base & Small & Tiny \\ 
    	\bottomrule
    	Number of encoder/decoder layers   & 6 & 6 & 4 \\ 
    	Input/Output layer dimension & 512 & 512 & 128 \\ 
    	Inner feed-forward layer dimension & 2048 & 1024 & 256 \\ 
    	Number of attention heads & 8 & 4 & 4 \\
    	\bottomrule
    \end{tabular}
}

    \caption{Model configurations for \textit{Base}, \textit{Small}, and \textit{Tiny}.}
    \label{tab:model_sizes}
\end{table}

%In our preliminary study, we found that even a \textit{Small} configuration, which is commonly used for low-resourced translation~\citep{zhu2020incorporating}, can still overfit on \dataset. 
%We use Adam with $\beta_1=0.9$, $\beta_2=0.98$ for model optimization. We start training with a warm-up phase (2,000 steps) where we linearly increase the learning rate from $10^{-7}$ to 0.005. Thereafter we decay the learning rate proportional to the number of updates. Each training batch contains at most 4,096 source/target tokens. We set label smoothing weight to 0.1, dropout to 0.3. We follow \cite{zhang2020neural} to early-stop the training if validation loss does not improve for ten epochs. We average the last ten checkpoints for inference as in \cite{vaswani2017attention} and \cite{wu2018pay}. We perform beam search with beam size set to 5. We report 4-gram BLEU and METEOR scores for all test sets. 
%\textmd{Multi-bleu.perl} was used to compute 4-gram BLEU scores for all test sets. 
%All models are trained and evaluated on one single machine with two Titan P100 GPUs. 

\section{Results on VaTex}
\label{app:vatex}
VaTex is a video-based MMT corpus that contains 129,955 English-Chinese sentence pairs for training, 15,000 sentence pairs for validation, and 30,000 sentence pairs for testing. Each pair of sentences is associated with a video clip. Since the testing set is not publicly available, we use half of the validation set for validating and the other half for testing. We apply the byte pair encoding algorithm on the lower-cased English sentences and split Chinese sentences into sequences of characters, resulting in a vocabulary of 17,216 English tokens and 3,384 Chinese tokens. We use the video features provided along with the VaTex dataset, in which each video is represented as $\mathbb{R}^{k*1024}$, where $k$ is the number of segments. Since some MMT systems take a ``global'' visual feature as input, we use 3D-Max-Pooling to extract the pooled representation $\mathbb{R}^{1024}$ for each video.

\begin{table}[htb!]
    \centering
    \resizebox{\linewidth}{!}{
    \begin{tabular}{l|c|c}
    \toprule
    \textbf{Model} & \textbf{BLEU} & \textbf{METEOR} \\ 
    \midrule
    Transformer & 35.82 & 59.02 \\
    $\quad$ +weight decay 0.1 & 36.32 & 59.38 \\
    $\quad$ +weight decay 0.01 & 36.07 & 59.14 \\
    $\quad$ +weight decay 0.001 & 35.92 & 59.22 \\
    \midrule
    Doubly-ATT & 36.05 (35.46) & 59.26 (58.84) \\ 
    Imagination & 36.25 (36.10) & 59.26 (59.15) \\
    Gated Fusion & 36.06 (36.01) & 59.34 (59.33) \\
    \model  & 36.35 (36.43) & 59.44 (59.57) \\ \hline
    % \model-att & 36.21 & 59.32 \\ 
    \bottomrule
    \end{tabular}
    }
    \caption{Results on VaTex En-Zh translation. Numbers in parentheses are the performance of the same model with random noise initialization.}
    \label{tab:vatex}
\end{table}

The results are shown in Table~\ref{tab:vatex}. We observe that although most MMT systems show improvement over the Transformer baseline, the gains are quite marginal. Indicating that although image-based MMT models can be directly applied to video-based MMT, there is still room for improvement due to the challenge of video understanding. We also note that (a) regularize the text-only Transformer with weight decay demonstrates similar gains as injecting video information into the models; (b) replacing video features with random noise replicate comparable performance, which further supports our findings in Section~\ref{sec:revisit}.

\section{Results on METEOR}
\label{app:meteor}

\begin{figure*}[]
    \centering
    \begin{minipage}{0.33\linewidth}
    \subfigure[Test2016.]{
    \includegraphics[width=\linewidth]{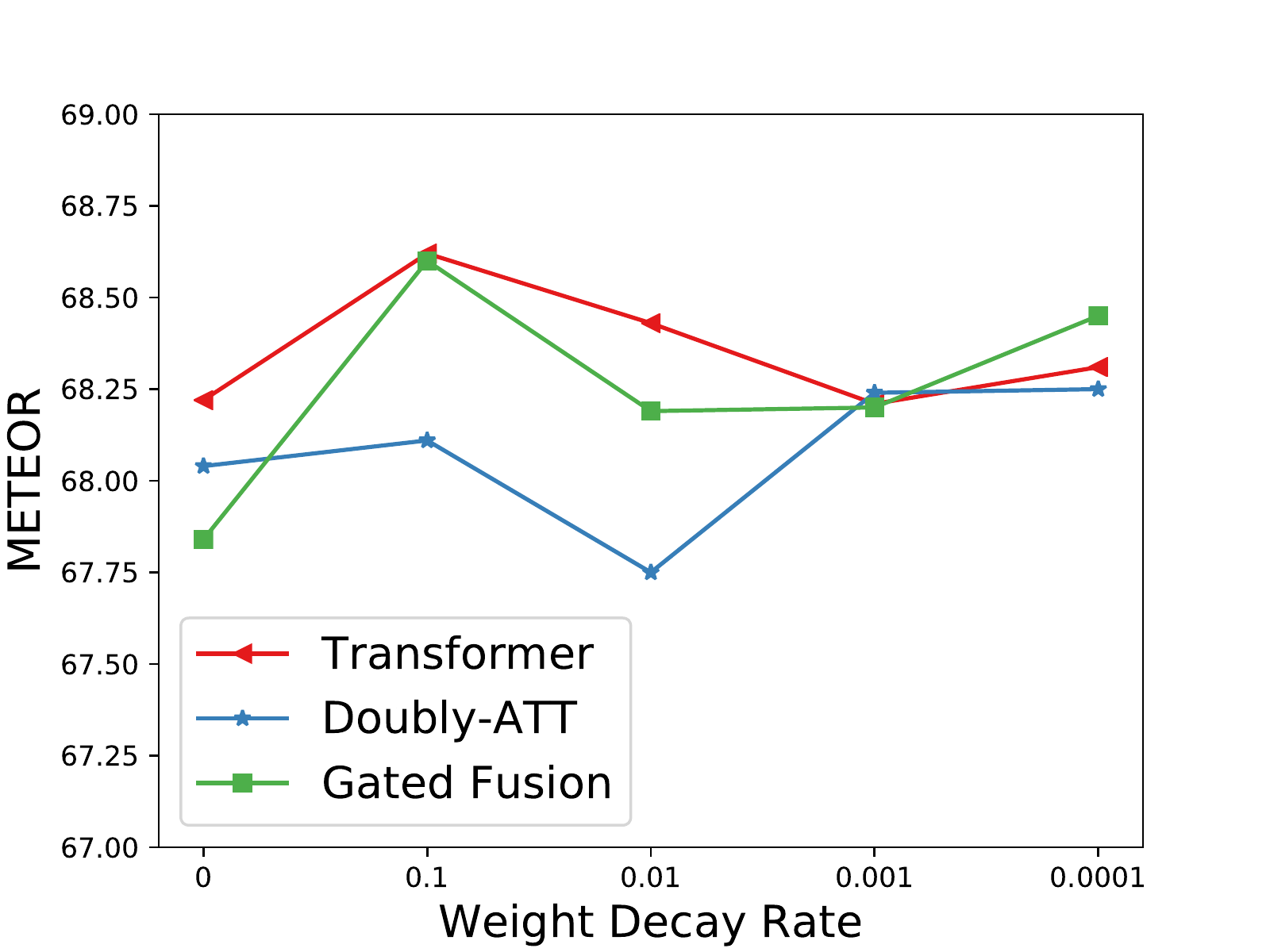}
    }
    \end{minipage}%
    \begin{minipage}{0.33\linewidth}
    \subfigure[Test2017.]{
    \includegraphics[width=\linewidth]{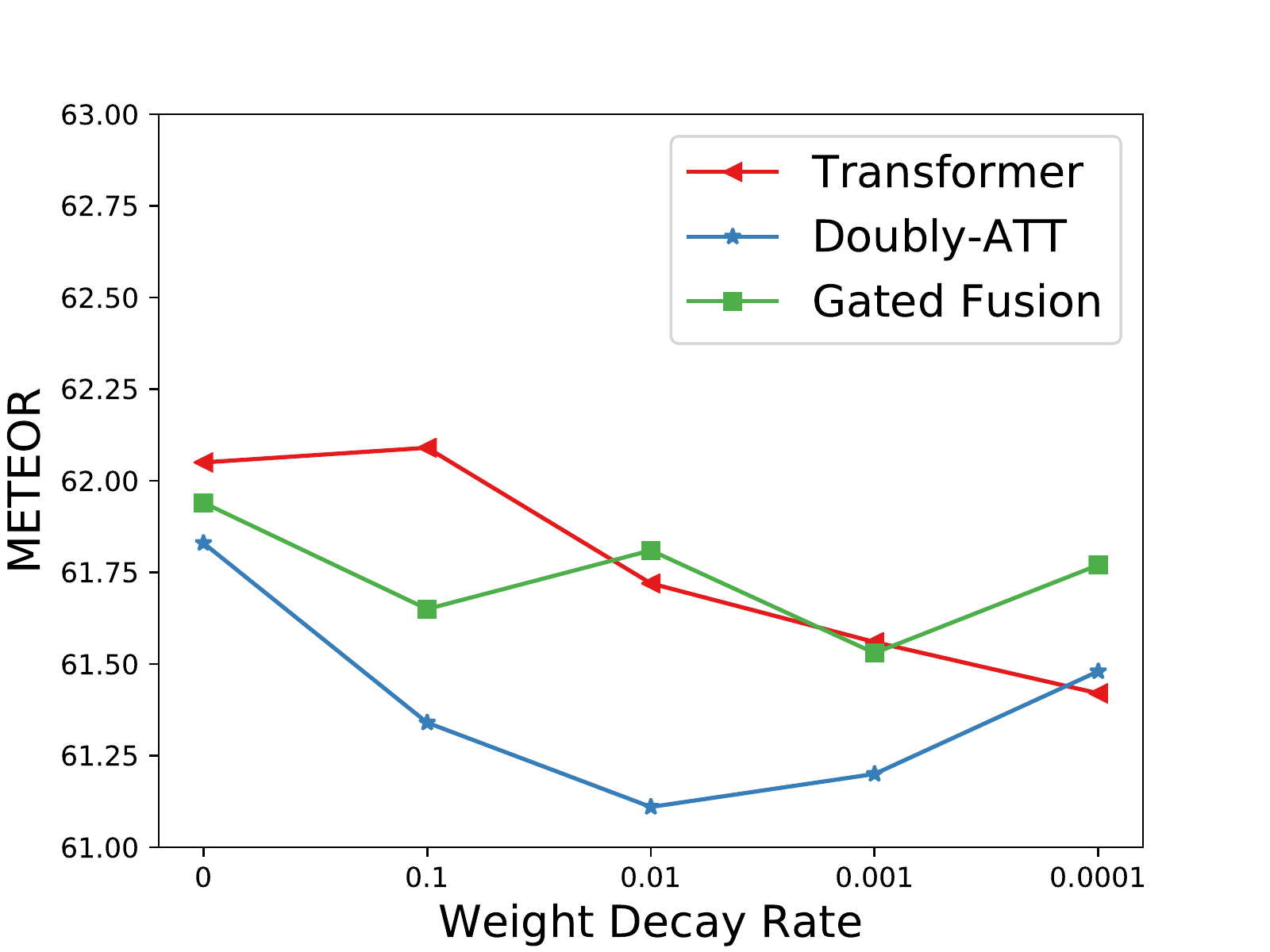}
    }
    \end{minipage}%
    \begin{minipage}{0.33\linewidth}
    \subfigure[MSCOCO.]{
    \includegraphics[width=\linewidth]{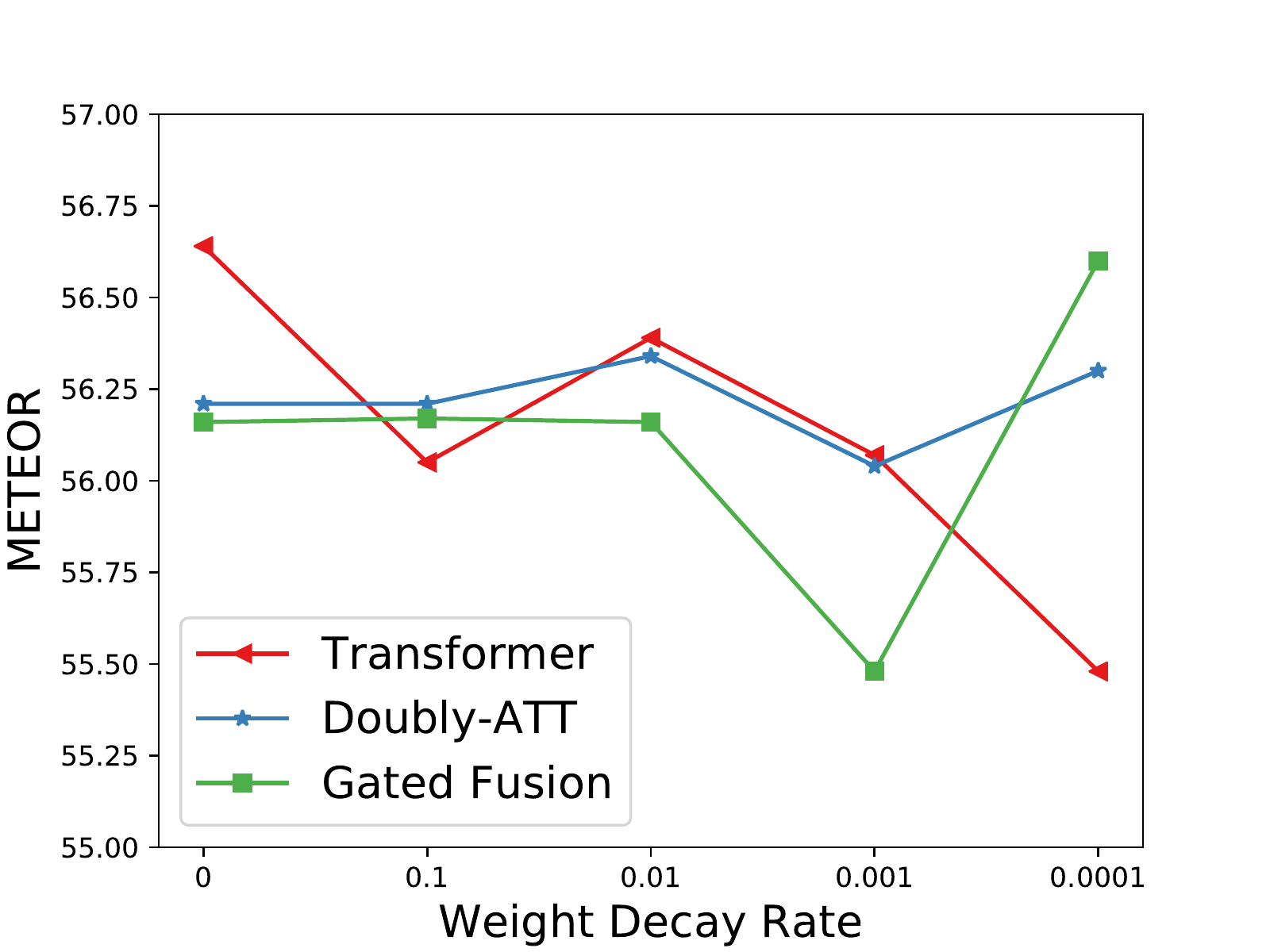}
    }
    \end{minipage}
    \caption{METEOR score curves on En$\to$De translation with different weight decay rate.}
    \label{fig:weight_decay_meteor}
\end{figure*}

\begin{table*}[]
    \centering
    \resizebox{\textwidth}{!}{
    \begin{tabular}{c|l|c|ccc|c|ccc}
    \toprule
    \multirow{2}{*}{\#} &\multirow{2}{*}{\textbf{Model}} & \multicolumn{4}{c|}{\textbf{En$\to$De}} & \multicolumn{4}{c}{\textbf{En$\to$Fr}} \\ 
    & & \textbf{\#Params} & \textbf{Test2016} & \textbf{Test2017} & \textbf{MSCOCO} &  \textbf{\#Params} &\textbf{Test2016} & \textbf{Test2017} & \textbf{MSCOCO} \\ 
    \midrule
    \multicolumn{10}{c}{\textit{Text-only Transformer}} \\ \hline
    1 &\textbf{Transformer}-Base & 49.1M & 65.92 & 60.02 & 54.73 & 49.0M & 80.09 & 74.93 & 68.57 \\
    2 &\textbf{Transformer}-Small & 36.5M & 66.01 & 60.80 & 55.95 & 36.4M & 80.71 & 75.74 & 69.10 \\
    3 &\textbf{Transformer}-Tiny & 2.6M  & 68.22 & 62.05 & 56.64 & 2.6M & 81.02 &75.62 & 69.43 \\ \hline
    \multicolumn{10}{c}{\textit{Existing MMT Systems}} \\ \hline
    4 &\textbf{GMNMT}$^\spadesuit$ & 4.0M & 57.6  & 51.9 & 47.6 & - & 74.9 & 69.3 & - \\
    5 &\textbf{DCCN}$^\spadesuit$ & 17.1M & 56.8  & 49.9 & 45.7 & 16.9M & 76.4  & 70.3 & 65.0 \\ 
    6 &\textbf{Doubly-ATT}$^\spadesuit$ & 3.2M & 68.04 & 61.83 & 56.21 & 3.2M & 81.12 & 75.71 & 70.25 \\
    7 &\textbf{Imagination}$^\spadesuit$ & 7.0M & 68.06 & 61.29 & 56.57 & 6.9M & 81.2  & 76.03 & 70.35 \\ \hline
    % 8 &\textbf{UVR-NMT}$^\diamondsuit$ & 2.9M & 40.79 & 32.16 & 29.02 & 2.9M & 61.00 & 53.20 & 43.71 \\ \hline
    \multicolumn{10}{c}{\textit{Our MMT Systems}} \\ \hline
    % 9 & \textbf{Balanced Fusion}$^\spadesuit$ & 2.8M & 66.64 & 58.62 & 54.12 & 2.8M & 78.16 & 72.90 & 67.71 \\
    9 &\textbf{Gated Fusion}$^\spadesuit$ & 2.9M & 67.84 & 61.94 & 56.15  & 2.8M & 80.97 & 76.34 & 70.51 \\
    10 &\textbf{\model}$^\diamondsuit$ & 2.9M & 67.97 & 61.71 & 56.33 & 2.9M & 81.29 & 76.09 & 70.24 \\
    % 12 &\textbf{\model-att}$^\diamondsuit$ & 3.3M & 67.55 & 61.66 & 56.25 & 3.2M & 80.71 & 75.68 & 70.15 \\
    \bottomrule
    \end{tabular}
    }
    \caption{METEOR scores on \dataset. Results in row 4 and 5 are taken from the original papers. $\spadesuit$ indicates conventional MMT models, while $\diamondsuit$ refers to retrieval-based models. Without further specification, all our implementations are based on the \textit{Tiny} configuration.}
    \label{tab:meteor}
\end{table*}
We also report our results based on METEOR~\citep{banerjee2005meteor}, which consistently demonstrates higher correlation with human judgments than BLEU does in independent evaluations such as in EMNLP WMT 2011~\footnote{http://statmt.org/wmt11/papers.html}. 
From Table~\ref{tab:meteor}, we can see that on En-Fr translation, MMT systems demonstrate similar improvements over text-only baselines in both METEOR and BLEU(see Table~\ref{tab:multi30k}). On En-De translation, however, MMT systems are mostly on-par with Transformer-tiny on METEOR and do not show consistent gains as BLEU. We hypothesis the reason being that En-De sets are created in a \textit{image-blind} fashion, in which the crowd-sourcing workers produce translations without seeing the images~\citep{frank2018assessing}. Such that source sentence can already provide sufficient context for translation. When creating the En-Fr corpus, the image-blind issue is fixed~\citep{mmt2017}, thus images are perceived as ``needed'' in the translation for whatever reason. Although BLEU is unable to elicit this difference, evaluation based on METEOR captured it and confirmed previous research. 
We also compute METEOR scores for our experiments that regularize models with random noise (see Table~\ref{tab:meteor_random}) and weight decay (see Figure~\ref{fig:weight_decay_meteor}). The results are consistent with those evaluated using BLEU and further complement our early findings.

\begin{table*}[]
    \centering
    \resizebox{\textwidth}{!}{
    \begin{tabular}{c|l|ccc|ccc}
    \toprule
    \multirow{2}{*}{\#} &\multirow{2}{*}{\textbf{Model}} & \multicolumn{3}{c|}{\textbf{En$\to$De}} & \multicolumn{3}{c}{\textbf{En$\to$Fr}} \\ 
    & &\textbf{Test2016} & \textbf{Test2017} & \textbf{MSCOCO} &\textbf{Test2016} & \textbf{Test2017} & \textbf{MSCOCO} \\ 
    \toprule
    1 &\textbf{Transformer} & 68.22 & 62.05 & 56.64 & 81.02 &75.62 & 69.43 \\ \hline
    2 &\textbf{Doubly-ATT} & 68.39(+0.35) & 61.83(+0.0) & 56.46(+0.25) & 81.27(+0.15) & 76.22(+0.51) & 70.21(-0.04) \\
    3 &\textbf{Imagination} & 67.93(-0.13) & 61.84(+0.55) & 56.49(-0.08) & 80.75(-0.45) & 76.57(+0.54) & 69.88(-0.47)\\
    4 &\textbf{Gated Fusion} & 68.25(+0.41) & 61.5(-0.44) & 55.93(-0.22) & 81.22(+0.25) & 76.01(-0.33) & 70.33(-0.18) \\
    % 5 &\model-Static & 41.60 & 32.98 & 29.54 &  &  &  \\
    \bottomrule
    \end{tabular}
    }
    \caption{METEOR scores on \dataset\ with randomly initialized visual representation. Numbers in parentheses indicate the relative improvement/deterioration compared with the original model with ResNet features.}
    \label{tab:meteor_random}
\end{table*}

%From Table~\ref{tab:meteor_random}, we observe that Doubly-ATT equipped with random noise (row 2) is consistently on-par with  (En$\to$De) or better than (En$\to$Fr) text-only baseline (row 1). On Doubly-ATT (En$\to$De), using random noise can even outperform the same model equipped with ResNet features. These adversarial evaluation complement our early findings illustrated using BLEU. 

%We can see from Figures~\ref{fig:reg_METEOR} (a) and (b) that with a large weight decay penalty (0.1), a text-only Transformer can also exhibit METEOR scores that are comparable to or even higher than MMT systems. These results are mostly consistent with those evaluated using BLEU (see Figure~\ref{fig:reg}).

\section{Results on IWSLT'14}
% \begin{wraptable}{r}{5.5cm}
\begin{table}[]
    \small
    \centering
    \begin{tabular}{lc}
    \toprule
    \textbf{Model} & \textbf{BLEU} \\ 
    \midrule
    Transformer-\textit{Small} & 28.62  \\
    $\quad$ +weight decay 0.0001 & 29.14 \\
    \model-\textit{Small}  & 29.03 \\
    %\model-\textit{Small} & 28.57 \\ \hline
    \bottomrule
    \end{tabular}
    \caption{BLEU score on IWSLT'14 EN$\to$DE translation.}
    \label{tab:iwslt-wmt}
\end{table}
\label{app:iwslt}
We also evaluate the retrieval-based model \model\ on text-only corpus --- IWSLT'14. The IWSLT'14 dataset contains 160k bilingual sentence pairs for En-De translation task. Following the common practice, we lowercase all words, split 7k sentence pairs from the training dataset for validation and concatenate \textit{dev2010}, \textit{dev2012}, \textit{tst2010}, \textit{tst2011}, \textit{tst2012} as the test set. The number of BPE operations is set to 20,000. We use the \textit{Small} configuration in all our experiments. The dropout and label smoothing rate are set to 0.3 and 0.1, respectively.  Since there is no images associated with IWSLT, we follow \cite{zhang2020neural} and retrieve top-5 images from Multi30K corpus.

From Table~\ref{tab:iwslt-wmt}, we see that Transformer without weight decay is marginally outperformed by \model, but achieves slightly higher BLEU scores when trained with a 0.0001 weight decay. Our discussion in Section~\ref{sec:revisit} sheds light on why visual context is helpful on non-grounded low-resourced datasets like IWSLT'14 --- for low-resourced dataset like IWSLT'14, injecting visual context help regularize model training and avoid overfitting.